\documentclass[10pt,twocolumn,letterpaper]{article}

\usepackage[pagenumbers]{cvpr} % To force page numbers, e.g. for an arXiv version

\usepackage{graphicx}
\usepackage{amsmath}
\usepackage{amssymb}
\usepackage{booktabs}
\usepackage{xcolor} 
\usepackage{kotex} % korean
\usepackage{rotating} % For comparison figure
\usepackage{multirow}
\usepackage{appendix}

\usepackage[accsupp]{axessibility}  % Improves PDF readability for those with disabilities.

\usepackage[pagebackref,breaklinks,colorlinks]{hyperref}

\usepackage[capitalize]{cleveref}
\crefname{section}{Sec.}{Secs.}
\Crefname{section}{Section}{Sections}
\Crefname{table}{Table}{Tables}
\crefname{table}{Tab.}{Tabs.}

\def\cvprPaperID{11711} % *** Enter the CVPR Paper ID here
\def\confName{CVPR}
\def\confYear{2023}

\begin{document}

\title{Diffusion Video Autoencoders: Toward Temporally Consistent \\
Face Video Editing via Disentangled Video Encoding}

\author{
Gyeongman Kim\textsuperscript{1} \quad Hajin Shim\textsuperscript{1} \quad  Hyunsu Kim\textsuperscript{2} \quad Yunjey Choi\textsuperscript{2} \quad Junho Kim\textsuperscript{2} \quad Eunho Yang\textsuperscript{1,3}\vspace{0.05in}\\
\textsuperscript{1}Korea Advanced Institute of Science and Technology (KAIST), South Korea\vspace{0.02in}\\
\textsuperscript{2}NAVER AI Lab \quad \textsuperscript{3}AITRICS, South Korea\\
{\tt\small\{gmkim, shimazing, eunhoy\}@kaist.ac.kr \quad \{hyunsu1125.kim, yunjey.choi, jhkim.ai\}@navercorp.com}
}

% \author{Gyeongman Kim\\
% KAIST\\
% gmkim@kaist.ac.kr
% \and
% Hajin Shim\\
% KAIST\\
% shimazing@kaist.ac.kr
% \and
% Naver\\
% Naver\\
% Naver
% \and
% Eunho Yang\\
% KAIST, AITRICS\\
% eunhoy@kaist.ac.kr
% }

\twocolumn[{
\maketitle
\begin{center}
    \vspace{-0.25in}
    \captionsetup{type=figure,width=0.94\textwidth}
    \includegraphics[width=0.94\textwidth]{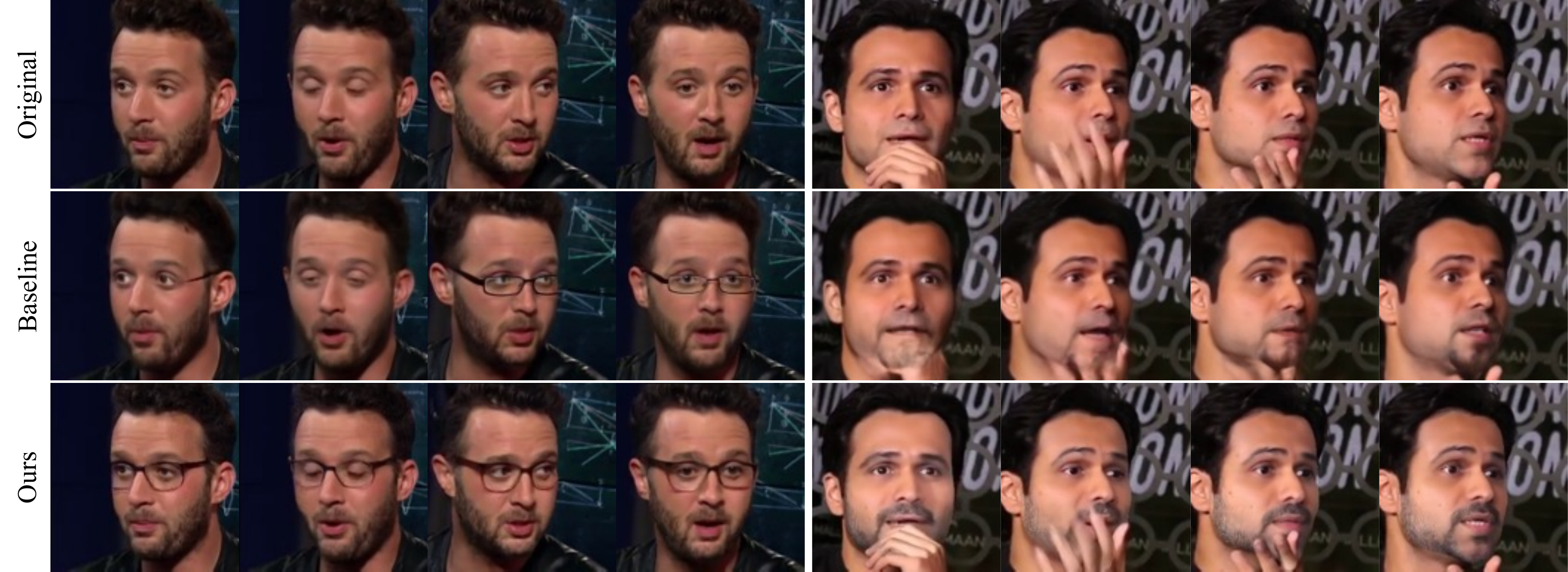}
    \vspace{-0.1in}
    \captionof{figure}{\textbf{Face video editing}. Our editing method shows improvement compared to the baseline \cite{stit} in terms of temporal consistency (left, ``eyeglasses'') and robustness to the unusual case such as the hand-occluded face (right, ``beard'').}
    \label{fig:main-result}
    %\vspace{0.05in}
\end{center}
}]

%\maketitle
% \begin{figure*}[tb]
%   \centering
%   \includegraphics[width=0.9\linewidth]{figures/Picture10.pdf}
%   \caption{Diffusion Video Autoencoders. We propose a novel human face video editing method for temporally coherent editing. }
%   \label{fig:main-result}
% \end{figure*}

%%%%%%%%% ABSTRACT
\begin{abstract}
\vspace{-0.1in}
   Inspired by the impressive performance of recent face image editing methods, several studies have been  naturally proposed to extend these methods to the face video editing task. One of the main challenges here is temporal consistency among edited frames, which is still unresolved.
   %Video editing must take into consideration not only the fidelity of each edited frame but also the temporal consistency between them. %Specifically, we aim to edit attributes of human face video in consideration of temporal consistency.
   To this end, we propose a novel face video editing framework based on diffusion autoencoders that can successfully extract the decomposed features - for the first time as a face video editing model - of identity and motion from a given video. This modeling allows us to edit the video by simply manipulating the temporally invariant feature to the desired direction for the consistency. 
   %decompose . We first encode the given video by decomposing it into  %decomposed feature with the assumption that the face video consists of a single
   %a single temporally invariant information (\eg identity) and per-frame  variant information (\eg motion and background). This modeling allows us to edit the video by simply decoding the decomposed features after manipulating the temporally invariant feature to the desired direction. 
   Another unique strength of our model is that, since our model is based on diffusion models, it can satisfy both reconstruction and edit capabilities at the same time, and is robust to corner cases in wild face videos (\eg occluded faces) unlike the existing GAN-based methods.\footnote{Project page: \href{https://diff-video-ae.github.io}{https://diff-video-ae.github.io}}
   %\B{Please visit our page: \href{https://diff-video-ae.github.io}{https://Diff-Video-AE.github.io/}}
   
   %For realistic video editing results, the decomposed features should be perfectly decoded as an original video and also the editability of the latent feature should be guaranteed. %an original video must be perfectly reconstructed through the decoding process with the decomposed features and each frame must be edited natuarally. 
   %As a way of doing so, we devise \emph{Diffusion Video Autoencoders} which are based on diffusion autoencoders to take the advantages of image reconstruction capability and editable semantic representation space and extend them for the video domain.
   %In our framework, we show that the features are properly decomposed into the temporally invariant and variant features and the video can be completely reconstructed through these features. In addition, we demonstrate through quantitative evaluation and qualitative analysis that our method preserves the temporal consistency better than the baselines by modifying a single static feature.
   %Finally, we show that our framework has no limitations such as not being able to edit the edge case (\eg covered face) due to the inversion ability of existing GAN-based face video editing methods.
\end{abstract}
\vspace{-0.21in}
%%%%%%%%% BODY TEXT
\section{Introduction}
\label{sec:intro}

%\iffalse
%\begin{comment}
%- Video editing 연구 배경

%Recently, as Generative Adversarial Networks (GANs) have shown remarkable performance in generating and editing images , further studies have been proposed to extend them to the time axis for videos. Although there is an additional process, each frame of video has been edited first by off-the-shelf image editing techniques independently until now. In particular, the StyleGAN-based method was used to gain strength in image generation and editing of face domain by utilizing the disentanglement property of semantic presentation space.
%\TODO{
%face  editing 하는게 중요한 문제다 -> 기존 face editing approach 소개(GAN, DPMs based) -> Video로 확장의 필요성(여기서 다루고자 하는 task 명확히 설명) -> 기존 GAN-based video editing 연구들의 접근방식 -> 한계점 1: inversion으로 edge case안되는 GAN의 한계 계승된다 -> 한계점 2: consistency 부분 문제점 -> 한계점1 대응: 그래서 우리는 diffusion model을 video editing에 처음 사용했다 강조, 최근 diffae같이 semantic representation을 사용할 수 있는 diffusion이 제안되면서 inversion에 있어 강점을 가져 edge case도 해결하고, editing도 가능하다 -> 한계점 2: 그러나, 기존의 diffusion-based image editing을 적용하더라고 frame별로 수정한다는 점에서 consistency 문제는 여전하다. 따라서, 우리는 video autoencoding에서의 decompose 를 통해 해결하고자 한다.
%우리 방식 설명 (모델 및 학습 및 manipulation)}

%\TODO{video로 확장의 필요성 및 기존연구들의 한계점. frame 단위로 접근 -> 2번째 문단 (temporal consistency)으로 연결} 

As one of the standard tasks in computer vision to change various face attributes such as hair color, gender, or glasses of a given face image, face editing has been continuously gaining attention due to its various applications and entertainment. In particular, with the improvement of analysis and manipulation techniques for recent Generative Adversarial Network (GAN) models  \cite{stylegan2, ganspace, interfacegan, sefa, styleclip}, we simply can do this task by manipulating a given image's latent feature. %instead of taking a picture again. 
%We can even make modifications we want in natural language form.
%Not only to edit  one of the element in the set of predefined attributes, we now can manipulate images in natural language format. 
In addition, very recently, many methods for face image editing also have been proposed based on Diffusion Probabilistic Model (DPM)-based methods that show  high-quality and flexible manipulation performance \cite{blended, sdedit, diffae, diffusionclip, dalle2, glide}.
%, and accordingly, significantly high-quality and flexible manipulation of images becomes achievable.

Naturally, 
%Recently, as image generation and manipulation models have shown remarkable performance \cite{stylegan2, interfacegan, ganspace, sefa, styleclip}, 
 further studies \cite{latent_transformer, stit, stylegan3video} have been proposed to extend image editing methods to incorporate the temporal axis for videos. Now, given real videos with a human face, these studies try to manipulate some target facial attributes with the other remaining features and motion intact.
 They all basically edit each frame of a video independently via off-the-shelf StyleGAN-based image editing techniques \cite{interfacegan, styleclip, latent_transformer}.%, with further engineering for each method. % for reconstruction \cite{pti} or seemless paste. % to consider temporal consistency. 
%In particular, the StyleGAN-based methods are usually taken to gain strength in 

%This is because StyleGAN has strength in image generation and editing of face images by utilizing the disentanglement property of the semantic representation space. 

Despite the advantages of StyleGAN in this task such as high-resolution image generation capability and highly disentangled semantic representation space, one harmful drawback of GAN-based editing methods is that the encoded real images cannot perfectly be recovered by the pretrained generator \cite{e4e, psp, restyle}. Especially, if a face in a given image is unusually decorated or occluded by some objects, the fixed generator cannot synthesize it. For perfect reconstruction, several methods \cite{pti, near_perfect_inv} are suggested to further tune the generator for GAN-inversion \cite{e4e, psp} on one or a few target images, which is computationally expensive. % because they require to additionally train the model for every sample, 
%Moreover, as expected, they can degrade the editability of GANs to some extent due to the distortion of the latent space.
Moreover, after the fine-tuning, the original editability of GANs cannot be guaranteed. This risk could be worse in video domain since we have to finetune the model on the multiple frames.% in the video.%should consider multiple frames in the video all at once. %

%However, they try to address temporal consistency implicitly with the incomplete assumptions such as smoothness in the latent space
%While several works,   address temporal inconsistency explicitly using optical flow  \cite{eccv22} or the dynamics of frames in the latent spaces \cite{ode}, they still exploiting the pretrained image generation models 

%- Video editing의 경우, Consistency 고려 필수 (기존 work 간단한 소개)

Aside from the reconstruction issue of existing GAN-based methods, it is critical in video editing tasks to consider the temporal consistency among edited frames to produce realistic results. 
%Unfortunately, it is difficult to expect temporal consistency by simply editing each frame of the video independently. This is true not only in the GAN-based methods but also in the DPM-based methods where the reconstruction is more complete. 
To address this, some prior works rely on the smoothness of the latent trajectory of the original frames \cite{stit} or smoothen the latent features directly \cite{stylegan3video} by simply taking the same editing step for all frames. However, smoothness does not ensure temporal consistency. Rather, the same editing step can make different results for different frames because it can be unintentionally entangled with irrelevant motion features. %struggle with consistently editing some attributes (\eg beard, eyeglasses) because the found manipulation directions for target attributes are entangled with other per-frame information.
For example, in the middle row of \cref{fig:main-result}, eyeglasses vary across time and sometimes diminish when the man closes his eyes. 
%Moreover, Xu \etal~\cite{eccv22} propose a method that refines each edited frame by measuring the degree of inconsistency between two frames explicitly by leveraging optical flow. However, measuring the degree of consistency with optical flow is still imperfect and there is a computational overhead to optimize all frames and the model in the editing process. 

%- GAN 기반 기존 work의 단점 가볍게 (pipeline, inversion에 큰 영향받음)
%In addition, prior video editing works have to invert all video frames to the latent space of StyleGAN first for using the StyleGAN-based off-the-shelf editing techniques~\cite{interfacegan, styleclip}. Although many approaches~\cite{pti, restyle, e4e, near_perfect_inv} have been proposed for realistic and editable GAN inversion, it is still difficult to invert the video frame of real video's various edge cases like covering the face with hands. Even if inversion is performed well, finding a reliable semantic direction on a fine-tuned latent space is hard for this edge case. \TODO{image editing with DM for video editing의 한계점}

% \TODO{문단 분리. 
% 1. 이 문제에diffusion model 처음으로 활용 (이전에는 stylegan 썼는데 왜 이걸로 넘어왔나?) ,
% 2. edit을 해야하니까 semantic diffae,
% 3. consistency issue를 위한 decompose,
% }
% \TODO{서술이 구체성이 떨어짐}

%- 우리가 제안하는 framework 소개

In this paper, we propose a novel video editing framework for human face video, termed \emph{Diffusion Video Autoencoder}, that resolves the limitations of the prior works. First, instead of GAN-based editing methods suffering from imperfect reconstruction quality, we newly introduce diffusion based model for face video editing tasks. As the recently proposed \emph{diffusion autoencoder} (DiffAE) \cite{diffae} does, thanks to the expressive latent spaces of the same size as input space, our model learns a semantically meaningful latent space that can perfectly recover the original image back and are directly editable. Not only that, for the first time as a video editing model, encode the decomposed features of the video: 1) identity feature shared by all frames, 2) feature of motion or facial expression in each frame, and 3) background feature that could not have high-level representation due to large variances. Then, for consistent editing, we simply manipulate a single invariant feature for the desired attribute (single editing operation per video), which is also computationally beneficial compared to the prior works that require editing the latent features of all frames.

We experimentally demonstrate that our model appropriately decomposes videos into time-invariant and per-frame variant features and can provide temporally consistent manipulation. Specifically, we explore two ways of manipulation. The first one is to edit features in the predefined set of attributes by moving semantic features to the target direction found by learning linear classifier in semantic representation space on annotated CelebA-HQ dataset \cite{progressiveGAN}. Additionally, we explore the text-based editing method that optimizes a time-invariant latent feature with CLIP loss \cite{stylegannada}. It is worth noting that since we cannot fully generate edited images for CLIP loss due to the computational cost, we propose the novel strategy that rather uses latent state of intermediate time step for the efficiency.

%appropriately
%The edited results demonstrate that our model conduct decomposition appropriately and manipulate consistent manipulation appropriately and outperform the baselines.
%\TODO{실험 결과에 대한 부분 보강}
% - diffae를 확장하여 video를 decoding했다  
% - 새로운 video editing frame 제안.  identity 수정해서 video 수정하게 했다
% - edge case를 잘 수행한다 (inversion 이 잘되니까 손으로 가려진 얼굴 등에 robust하게 잘 적용할 수 있음)
% - text 기반의 새로운 editing 방식 제안한다. CLIP-based의 어쩌구 이름 짓기 

To summarize, our contribution is four-fold:
\begin{itemize}

%\item We devise the Diffusion Video Autoencoders that extend the Diffusion Autoencoders~\cite{diffae} of images to encode a video to a one time-invariant representation and the other time-dependent features which can be reconstructed to the original video to a near-perfect level. 
\vspace{-0.05in}
\item We devise diffusion video autoencoders based on diffusion autoencoders~\cite{diffae} that decompose the video into a single time-invariant and per-frame time-variant features for temporally consistent editing. 
\vspace{-0.05in}
\item %We propose a novel human-face-video editing framework 
Based on the decomposed representation of diffusion video autoencoder, face video editing can be conducted by editing only the single time-invariant identity feature and decoding it together with the remaining original features. 
\vspace{-0.05in}
\item Owing to the nearly-perfect reconstruction ability of diffusion models, our framework can be utilized to edit exceptional cases such that a face is partially occluded by some objects as well as usual cases.

\vspace{-0.05in}

\item In addition to the existing predefined attributes editing method, we propose a text-based identity editing method based on the local directional CLIP loss \cite{clip, stylegannada} for the intermediately generated product of diffusion video autoencoders.

\end{itemize} 

\section{Related Work}

% \begin{figure}
%   \centering
%   \includegraphics[width=\linewidth]{figures/Picture5.pdf}
%   \caption{Failure cases  of existing per-frame editing methods~\cite{stit} (beard(above), eyeglasses(below)).}
%   \label{fig:failure-case}
% \end{figure}

\paragraph{Video editing}
When editing a given real video, it is essential to preserve temporal consistency.  First, Lai \etal \cite{blind} consider editing global features of videos such as artistic style transfer, colorization, image enhancement, etc. To encourage temporal consistency, they train sequential image translator with temporal loss defined as the warping error between the output frames. 

Different from this work, our target task is to edit facial attributes of human face video. In other words, we aim to change face-related attributes such as eyeglasses, beards, etc. For this task, Yao \etal \cite{latent_transformer} are the first that propose the editing pipeline: 1) align and crop the target face area, 2) encode these frames to latent features, manipulate and decode them, and 3) unalign and paste the edited frames to the original video. However, every step of this pipeline is conducted independently for each frame. %It is worth noting that the following video manipulation works covered in this section also take the variant of this pipeline.
For the manipulation step, they try to find disentangled editing directions for a desired attribute %by learning latent transformers
and take the same step for every frame, while expecting the disentanglement brings consistency automatically. However, the results show inconsistency and we conjecture that both latent space and learned direction are still entangled with many other attributes.
Next, %while following the similar pipeline,
Tzaban \etal~\cite{stit} additionally fine-tune the pretrained StyleGAN \cite{pti} to enhance reconstructability. Meanwhile, they assume that temporal consistency would be preserved when applying the same editing step to the latent features of the original frames. However, their assumption is not always true for all attributes, especially for beards and eyeglasses as shown in \cref{fig:main-result}. Alauf \etal \cite{stylegan3video} also try to avoid temporal inconsistency by smoothing latent features.

While the methods in the previous paragraph handle inconsistency implicitly, several works\cite{eccv22, ode} try to solve it more directly. After per-frame editing, Xu \etal \cite{eccv22} further optimize latent codes and the pretrained StyleGAN to enhance consistency between the frame pairs defined by involving bi-directional optical flow. On the other hand, Xia \etal~\cite{ode} propose to learn the dynamics of inverted GAN latent codes of video frames. The learned dynamics are used to generate the subsequent latent features after editing only the first frame. %All the above methods are based on the pretrained StyleGAN.

Additionally, although Skorokhodov \etal~\cite{stylegan_v} mainly target to generate video, they also demonstrate manipulation results of the generated video. They model the video with a time-invariant content code and per-frame motion codes. Based on this modeling, the generated video can be manipulated with a target text by optimizing its content vector with CLIP loss. However, they cannot edit the real videos due to the absence of an encoding method. In contrast, our proposed method is capable of encoding and manipulating realistic face videos. 

\vspace{-0.1in}

\paragraph{Diffusion models} 
%Recently, diffusion probabilistic models (DPMs) is one of the most prominent generative model \cite{ddpm, improved_ddpm, ddim, dalle2}. 
Denoising diffusion probabilistic models (DDPMs) \cite{ddpm}
associate image generation with the sequential denoising process of isotropic Gaussian noise. The model is trained to predict the noise from the input image. Unlike other generative models such as GANs and most traditional-style VAEs that encode input data in a low-dimensional space, diffusion models have a latent space that is the same size as the input. Although DDPMs require a lot of feed-forward steps to generate samples, their image fidelity and diversity are superior to other types of generative models. 
Compared to DDPMs that assume a Markovian noise-injecting forward diffusion process, Denoising diffusion implicit models (DDIMs) \cite{ddim} assume a non-Markovian forward process that has the same marginal distribution as DDPMs, and use its corresponding reverse denoising process for sampling, which enables acceleration of the rather onerous sampling process of DDPMs. 
DDIMs also utilize a deterministic forward-backward process and therefore show nearly-perfect reconstruction ability, which is not the case for DDPMs. In this paper, we adopt conditional DDIMs to encode, manipulate and decode real videos. 

\vspace{-0.1in}

\paragraph{Image editing with diffusion models}
There are various attempts to manipulate images with diffusion models \cite{diffae, diffusionclip, sdedit, blended, dalle2, glide} such as text-guided inpainting \cite{glide,dalle2,blended}, stroke-based editing \cite{sdedit}, and style transfer \cite{diffusionclip}. Among those, we are especially interested in facial attribute editing tasks. Kim \etal~\cite{diffusionclip} propose text-based image manipulation by optimizing an unconditional diffusion model to minimize local directional CLIP loss \cite{stylegannada}. This method is superior to GAN-based methods as it can successfully edit even occluded or overly decorated faces by the excellent reconstruction ability of diffusion models. However, due to the absence of semantically meaningful latent space, there exist some semantic features that are unable to be changed. Another parallel work proposes diffusion autoencoder (DiffAE) \cite{diffae}, which takes a learnable encoder to obtain semantic representations of which the underlying diffusion model is conditioned. The model can perform face attribute manipulation by moving the semantic vector to a target direction. Based on DiffAEs, we design our diffusion video autoencoders to perform temporally-consistent video editing.

\begin{figure*}
  \centering
  \begin{subfigure}{0.40\linewidth}
    \includegraphics[width=\linewidth]{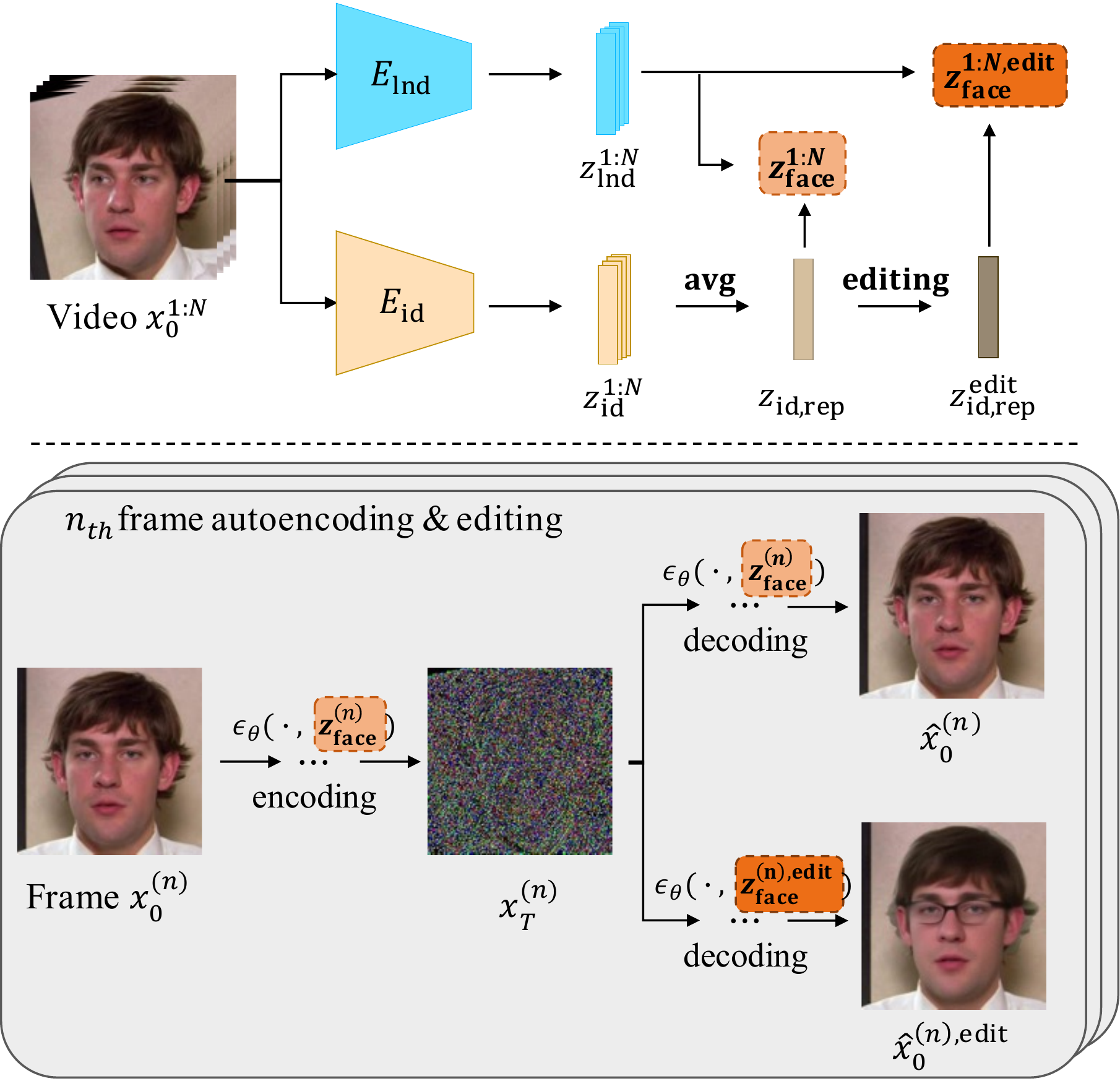}
    \caption{Video autoencoding and editing pipeline.}
    \label{fig:dva-a}
  \end{subfigure}
  \hfill
  \begin{subfigure}{0.56\linewidth}
    \includegraphics[width=\linewidth]{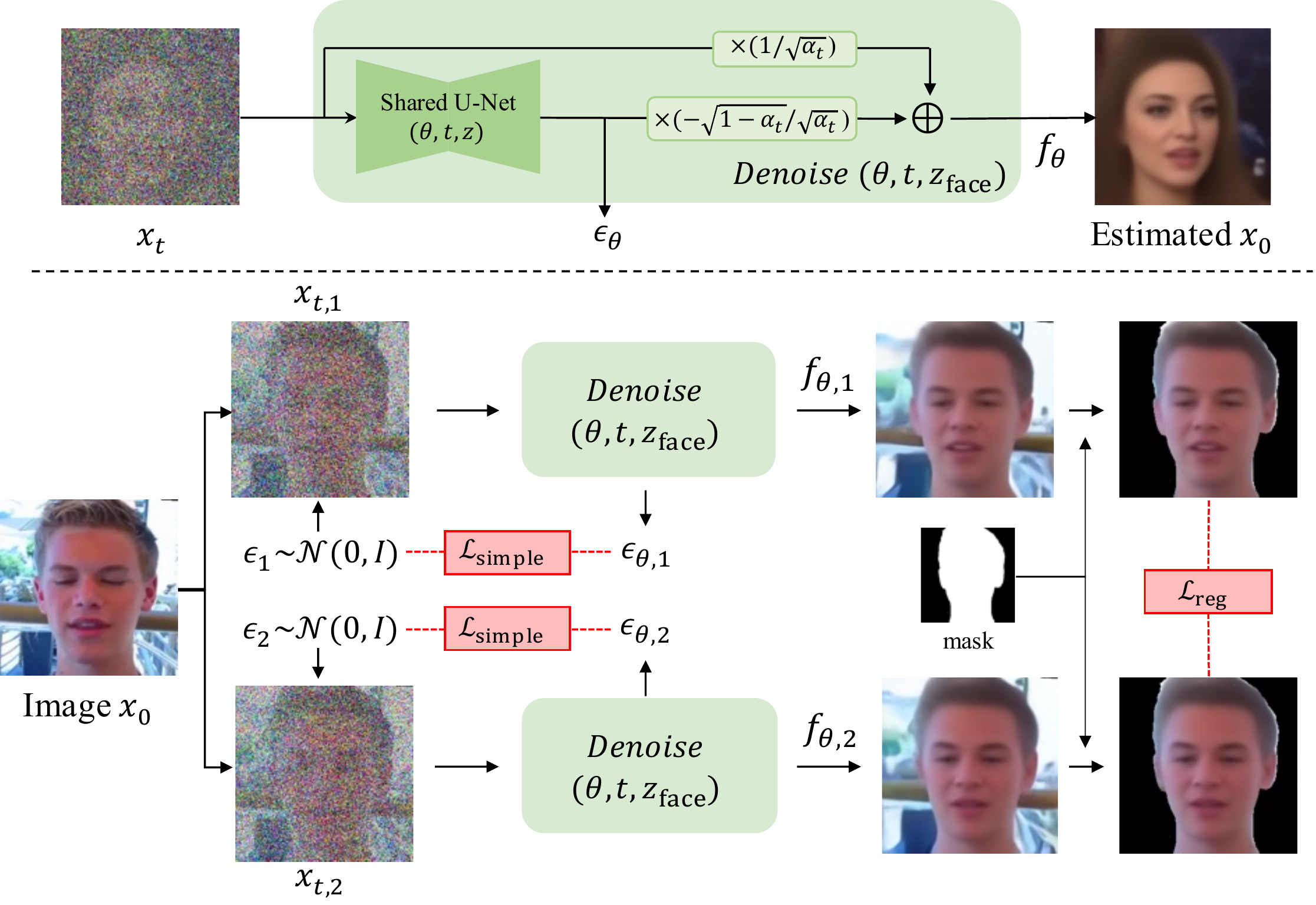}    
    \caption{Training objective.}
    \label{fig:dva-b}
  \end{subfigure}
  \vspace{-0.05in}
  \caption{Overview of our Diffusion Video Autoencoder.}
  \label{fig:dva}
  \vspace{-0.1in}
\end{figure*}

\section{Preliminaries}
\paragraph{Diffusion probabilistic models (DPMs)} 
\label{subsec:dm}
% DDPM
%Diffusion probabilistic models
DPMs~\cite{noneq_thermo, ddpm} are generative models that attempt to approximate the data distribution $q(x_0)$ via $p_\theta(x_0)$ from the reverse prediction of Markovian diffusion process $q(x_{1:T}|x_0) = \prod_{t=1}^T q(x_t|x_{t-1})$. Here, the forward process of DPM is a Gaussian noise perturbation $q(x_t|x_{t-1})=\mathcal{N}(\sqrt{1-\beta_t}x_{t-1},\beta_tI)$ where $\beta_t$ is fixed or learned variance schedule with increasing $\beta_1<\beta_2<\cdots<\beta_T$, adding gradually increasing noise to data $x_0$, and its reverse process is denoising of $x_t$ step by step in reverse order. %Since the forward process is a continuous form of Gaussian transition $q(x_t|x_{t-1})=\mathcal{N}(\sqrt{1-\beta_t}x_{t-1},\beta_tI)$, 
From this definition, a noisy image $x_t$ at step $t$ can be expressed as $x_t=\sqrt{\alpha_t}x_0+\sqrt{1-\alpha_t}\epsilon,~
\epsilon\sim\mathcal{N}(0,I)$
 where $\alpha_t=\prod_{s=1}^{t} (1-\beta_s)$. While the true reverse process $q(x_{t-1}|x_t)$ is intractable, %requires the entire dataset to be computed, %, but if $\beta_t$ is small enough, this transition also become a Gaussian. However, 
 %Since it is difficult to figure out data distribution $q(x_0)$ exactly, 
 diffusion models %estimate the reverse process by
 approximate $q(x_{t-1}|x_t,x_0)$ with $p_\theta(x_{t-1}|x_t)=\mathcal{N}(\mu_\theta(x_t,t),\sigma_t)$ by minimizing the variational lower bound of negative log-likelihood. % parameterized as $\mu_\theta(x_t,t) = \frac{1}{\sqrt{\alpha_t}}(x_t - \frac{\beta_t}{\sqrt{1-\alpha_t}}\epsilon_\theta(x_t,t))$. 
Finally, by reparameterizing $\mu_\theta$, the objective is given as % a variant of the variational lower bound of negative log-likelihood %and reparameterization trick~\cite{ddpm} are used
 %which is equal to %$$\mathcal{L}_{ddpm}=
 $\mathbb{E}_{x_0\sim q(x_0),\epsilon_t\sim\mathcal{N}(0,I),t} \begin{Vmatrix}\epsilon_\theta(x_t,t)-\epsilon_t\end{Vmatrix}_2^2$ where the parameterized model %$x_t=\sqrt{\alpha_t}x_0+\sqrt{1-\alpha_t}\epsilon_t$ and 
 $\epsilon_\theta(x_t,t)$ estimates the true Gaussian noise term $\frac{1}{\sqrt{1-\alpha_t}}(x_t - \sqrt{\alpha_t} x_0)$ in the forward process. 

% DDIM
Speeding up a time-consuming sampling process of diffusion models is one of the core research topics related to diffusion models \cite{improved_ddpm,ddim}. Among others, Song \etal \cite{ddim} propose DDIM, which assumes a non-Markovian forward process $q(x_t|x_{t-1},x_0)$ that has the same marginal distribution $q(x_t|x_0)$ with DDPMs. The corresponding generative process of DDIM is $x_{t-1}=\sqrt{\alpha_{t-1}}f_\theta(x_t,t)+\sqrt{1-\alpha_{t-1}-\sigma_t^2}\epsilon_\theta(x_t,t)+\sigma_t^2z$ where $z\sim\mathcal{N}(0,I)$, $\sigma$ is sampling stochasticity and $f_\theta(x_t,t)$, the estimated value of $x_0$ at time step $t$, is as follow:
$$f_\theta(x_t,t) = \frac{x_t - \sqrt{1-\alpha_t}\epsilon_\theta(x_t,t)}{\sqrt{\alpha_t}}.$$ 
When $\sigma$ is set to $0$, this process becomes deterministic. 
%\R{The deterministic forward process can be defined by treating the reverse sampling process as an ordinary differential equation (ODE) and inverse it.} 
With deterministic forward and reverse processes, we can obtain the latent code of each original sample, which can reconstruct the original sample through the reverse process.
\vspace{-0.1in}
%which is a characteristic of DDIM. Since the marginal distribution is the same as DDPM, DDIM also has the same objective as DDPM.

\paragraph{Diffusion autoencoders}
% DiffAE
Due to the determinacy of its sampling process, DDIM can reconstruct the original image $x_0$ from $x_T$ obtained through $T$ forward process steps. While $x_T$ can be considered as a latent state with the same size as $x_0$, it does not contain high-level semantic information~\cite{diffae}. To supplement this, Prechakul \etal propose diffusion autoencoder (DiffAE)~\cite{diffae} based on DDIM. DiffAE utilizes two forms of latent variables: $z_\texttt{sem}$ for the useful high-level semantic representation and $x_T$ for the remaining low-level stochasticity information. DiffAE introduces a semantic encoder $\texttt{Enc}(x_0)$ which extracts $z_\texttt{sem}$ from an image, and makes a noise estimator $\epsilon_\theta(x, t, z_\texttt{sem})$ conditioned on $z_\texttt{sem}$. Stochastic latent  $x_T$ is calculated through the deterministic DDIM forward process \cite{ddim} that involves the noise estimator  $\epsilon_\theta$ given $z_\texttt{sem}$. Then, $(z_\texttt{sem}, x_T)$ is decoded to %diffusion-based decoder that 
 reconstruct the corresponding original image $x_0$ % using the given $z_{sem}$ and $x_T$ 
 with $p_\theta(x_{0:T}|z_\texttt{sem})=p(x_T)\prod_{t=1}^T p_\theta(x_{t-1}|x_t,z_\texttt{sem})$. %DiffAE involves $z_{sem}$ as condition of DDIM through adaptive group normalization layer in the noise estimator of the U-Net structure. 
DiffAE is trained with simple DDPM loss, as done with DDIM.  
 %Similar to DDIMs, DiffAEs optimize the following 
 %Although use same DDIM sampling process, diffusion autoencoder's
 %objective: $$\mathcal{L}_{diffae}=\mathbb{E}_{x_0\sim q(x_0),\epsilon_t\sim\mathcal{N}(0,I),t}\begin{Vmatrix}\epsilon_\theta(x_t,t,z_{sem})-\epsilon_t\end{Vmatrix}_2^2.$$ 
 The differences between DiffAE and DDIM are that an additional variable %This is same objective as, but 
 $z_\texttt{sem}$ is involved as a conditioning variable during the reverse process (and thus the training process) and that the noise estimator $\epsilon_\theta$ and semantic encoder $\texttt{Enc}$ are jointly trained. As a result, through the training phase, $z_\texttt{sem}$ is learned 
 %at the same time, $z_{sem}$ representation space learn 
  to capture the high-level semantic information of the image, and the low-level stochastic variations % of the images 
   remain in $x_T$.

\section{Diffusion Video Autoencoders}
\label{sec:Diffusion Video Autoencoders}

%Inspired by the meaningful and editable representation space of DiffAEs, 
In this section, we introduce a video autoencoder, called \emph{Diffusion Video Autoencoder}, specially designed for face video editing to have 1) superb reconstruction performance and  2) an editable representation space for the identity feature of the video disentangled with the per-frame features changing over time. The details of the model components and the training procedure are explained in \cref{sec:Disentangled Video Encoding} and then two different editing methods based on our model are presented in \cref{sec:Video Editing Framework}. We provide the overview in \cref{fig:dva}. % contain the overview for better under.

%followed by the video editing framework using our proposed model in \cref{sec:Video Editing Framework}. 
%we introduce a video autoencoder designed specially for face video editing, called \emph{Diffusion Video Autoencoder} that satisfy  having semantically meaningful representation space. 
%For temporally consistent face video editing, we design autoencoders for face video, called Diffusion Video Autoencoders, to encode the video %through our diffusion video autoencoder into the high-level semantic space and the stochastic latent space in disentangled manner. In \cref{sec:Disentangled Video Encoding}, we explain how our diffusion video autoencoder decomposes and learns the video features. Then we introduce a video editing framework using the trained diffusion video autoencoder and a novel text-based editing method using CLIP guidance in \cref{sec:Video Editing Framework}.

\subsection{Disentangled Video Encoding}
\label{sec:Disentangled Video Encoding}

% Design choice (3 부분으로 나뉘어짐)
To encode the video with $N$ frames $\{x_{0}^{(n)}\}_{n=1}^N$, we consider the time-invariant feature of human face videos as human identity information, and the time-dependent feature for each frame as motion and background information. Among these three,  %while 
identity or motion information relevant to a face is appropriate to be projected to a low-dimensional space to extract high-level representation. %has general characteristics, 
Comparatively, the background shows high variance with arbitrary details and changes more with the head movement by cropping and aligning the face region.
Therefore, it could be very difficult to encode the background information into a high-level semantic space. % with low dimension.
Thus, identity and motion features %, which are relevant to face, 
are encoded in high-level semantic space $z_\texttt{face}^{(n)}$, combining identity feature $z_\texttt{id}$ of the video and motion feature $z_\texttt{lnd}^{(n)}$ of each frame, and the background feature is encoded in noise map $x_T^{(n)}$ (see \cref{fig:dva-a}). We denote $z_\texttt{id}$ without superscript $(n)$ for frame index since it is the time-invariant and shared across all frames of the video.

To achieve this decomposition, our model consists of two separated semantic encoders - an identity encoder $E_\texttt{id}$ and a landmark encoder $E_\texttt{lnd}$ -  and a conditional noise estimator $\epsilon_\theta$ for diffusion modeling. The encoded features $(z_\texttt{id}, z_\texttt{lnd}^{(n)})$ by the two encoders are concatenated and passed through an MLP to finally get a face-related feature $z_\texttt{face}^{(n)}$.
%Before covering the training process of our model, we first explain about the overall model components. To obtain $z_\texttt{face}$, 
%we use two separated encoders: an ID encoder $E_\texttt{id}$ for identity and a landmark encoder $E_\texttt{lnd}$ for motion and the encoded features by the two encoders are concatenated and passed through an MLP to finally get $z_\texttt{face}$. To be clear, through $E_\texttt{id}$, we want to extract the time-invariant identity feature $z_\texttt{id}$ which is not different from frame to frame and shared across all frames of the video. 
Next, to encode noise map $x_T^{(n)}$, we run the deterministic forward process of DDIM using noise estimator $\epsilon_\theta$ with conditioned on $z_\texttt{face}^{(n)}$.
Since noise map $x_T^{(n)}$ is a spatial variable %of a large dimension 
with the same size as the image, it is expected that information in the background can be encoded more easily without loss of spatial information. Then, the encoded features $(x_T^{(n)}, z_\texttt{face}^{(n)})$ can be reconstructed to the original frame by running generative reverse process of conditional DDIM in a deterministic manner:  
\vspace{-0.1in}
$$p_\theta(x_{0:T}|z_\texttt{face})=p(x_T)\prod_{t=1}^T p_\theta(x_{t-1}|x_t,z_\texttt{face}).$$ 
\vspace{-0.1in}
%$x_T$ contains the information that cannot be included in the semantic feature, which is expected to contain background information.

To obtain the identity feature disentangled with the motion, we choose to leverage a pretrained model for identity detection, named ArcFace \cite{arcface}. ArcFace is trained to classify human identity in face images regardless of poses or expressions, so we expect it to provide the disentangled property we need. 
Nevertheless, when an identity feature is extracted for each frame through the pretrained identity encoder, the feature may be slightly different for each frame because some frames may have partial identity features for some reasons (\eg excessive side view poses).
%each frame might have a slightly different or some frames have not enough identity features caused by excessive side view poses. 
To alleviate this issue, we average the identity features $z_\texttt{id}^{(n)} = E_\texttt{id}(x_0^{(n)})$ of all frames in the inference phase.  % feature. Therefore, we assume that we will extract very similar identity features from all frames regardless of the head pose by using the pretrained network for the face recognition task~\cite{idnet}.
%It further compensates for hidden ID information in some frames caused by excessive side view poses by averaging the ID features of all frames in inference phase. 
Similarly, we obtain per-frame motion information via a pretrained landmark detection model \cite{landmark} which outputs the position of face landmarks. %Obviously, there is also an option to train the semantic encoders using some techniques to encourage disentanglement~\cite{disentangled_lifespan, face_swap}, but these require to generate images to compute the loss. However, diffusion models take a lot of steps to synthesize images from latent features, which is an obstacle to adopting this option. 
%Furthermore, 
Several studies~\cite{face_decom_map, face-vid2vid} have %already 
shown that it is possible to have a sufficiently disentangled nature by extracting features through a pretrained encoder without learning. Therefore, diffusion video autoencoders extract an identity and also a landmark feature of the image through the pretrained encoders and map them together to the high-level semantic space for face features through an additional learnable MLP.

Next, we explain how the learnable part of our model is trained. %training process of our diffusion video autoencoders. 
\cref{fig:dva-b} summarizes our training process. For simplicity, from now on, we drop the superscript of frame index. Our objective consists of two parts. The first one is the simple version of DDPM loss  \cite{ddpm} as  $$\mathcal{L}_\texttt{simple}=\mathbb{E}_{x_0\sim q(x_0),\epsilon_t\sim\mathcal{N}(0,I),t} \begin{Vmatrix}\epsilon_\theta(x_t,t,z_\texttt{face})-\epsilon_t\end{Vmatrix}_1$$ 
where $z_\texttt{face}$ is an encoded high-level feature of input image $x_0$.
It encourages the useful information of the image to be well contained in the semantic latent $z_\texttt{face}$ and exploited by $\epsilon_\theta$ for denoising. Secondly, we devise a regularization loss to hinder face information (identity and motion) from leaking to $x_T$ but 
%$x_t$ to encode as much information about the face as possible in the
contained in $z_\texttt{face}$ as much as possible %and not in $x_T$
for clear decomposition between background and face information. 
If some face information is lost in %during the mapping process of 
$z_\texttt{face}$, the lost information would remain in the noise latent $x_T$ unintentionally. To avoid it, % and reduce the effect of $x_T$ for face features, 
we sample two different Gaussian noises $\epsilon_1$ and $\epsilon_2$ to obtain different noisy samples $x_{t,1}$ and $x_{t, 2}$, respectively. % in training phase, 
Then, we minimize the difference between the estimated original images $f_{\theta,1}$ and $f_{\theta, 2}$ except for the background part as: 
$$\mathcal{L}_\texttt{reg}=\mathbb{E}_{x_0\sim q(x_0),\epsilon_1,\epsilon_2\sim\mathcal{N}(0,I),t} \begin{Vmatrix}f_{\theta,1}\odot m-f_{\theta,2}\odot m\end{Vmatrix}_1$$ where $m$ is a segmentation mask of a face region in the original image $x_0$ and $f_{\theta,i}=f_{\theta}(x_{t,i},t,z_\texttt{face})$. 
As a result, the effect of noise in $x_t$ on the face region will be reduced and $z_\texttt{face}$ will be responsible for face features for the generation (See \cref{fig:ablation-reg}). 
%\R{In other words, the facial features are encouraged to be contained in the high-level semantic latent $z_\texttt{face}$ as much as possible.} 
In \cref{sec:Ablation study}, we demonstrate the desired effect of $\mathcal{L}_\texttt{reg}$. The final loss of diffusion video autoencoders is as follows: $\mathcal{L}_\texttt{dva} = \mathcal{L}_\texttt{simple} + \mathcal{L}_\texttt{reg}.$

\begin{figure}
  \centering
  %\begin{subfigure}{0.38\linewidth}
  %  \includegraphics[width=\linewidth]{figures/Picture3.pdf}
  %  \caption{Video editing framework of diffusion video autoenocder}
  %  \label{fig:editing-a}
  %\end{subfigure}
  %\hfill
  %\begin{subfigure}{0.56\linewidth}
    \includegraphics[width=\linewidth]{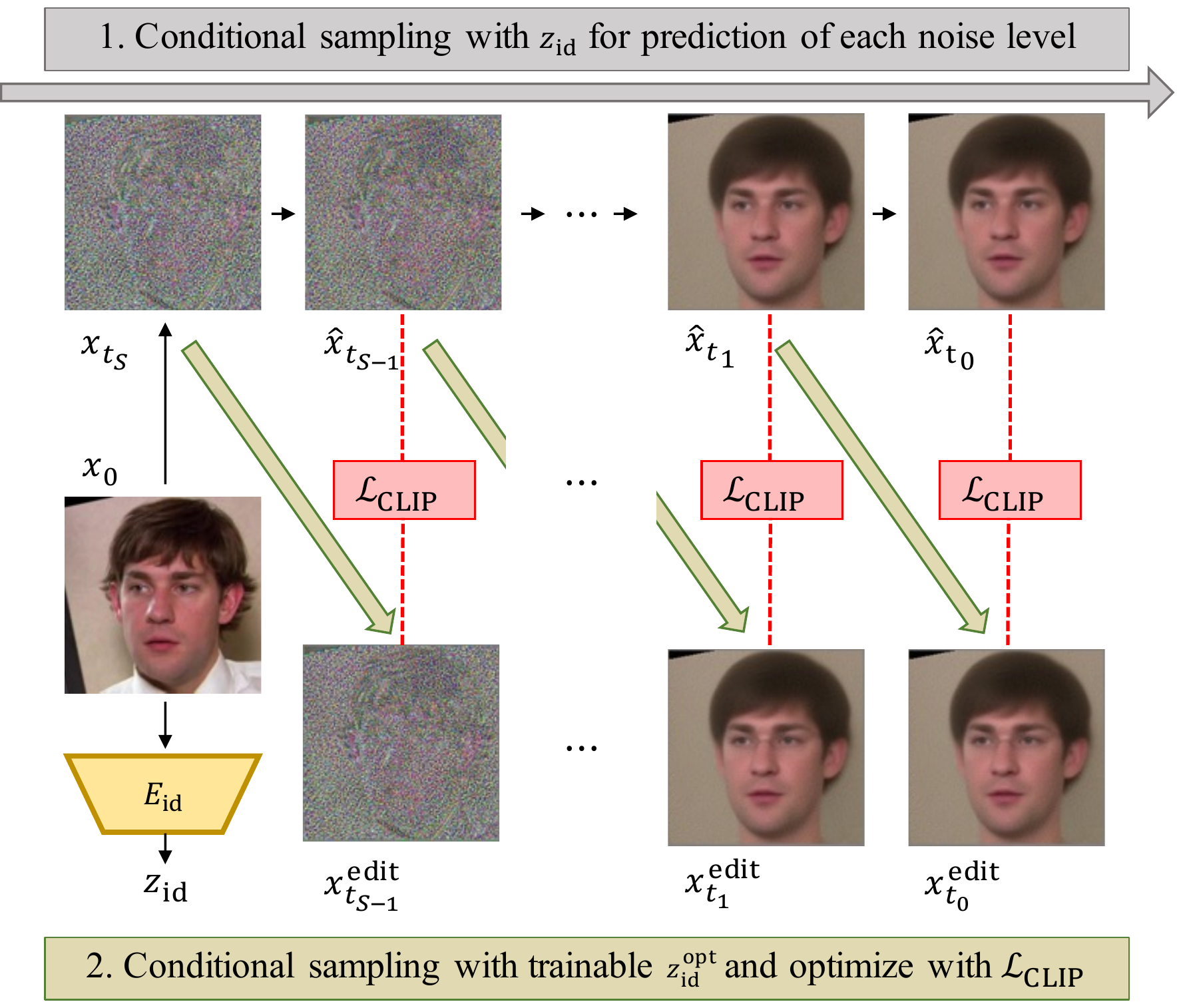}  
    \vspace{-0.2in}
    \caption{Process of computing noisy CLIP loss.}
    \label{fig:editing}
    \vspace{-0.2in}
  %\end{subfigure}
  %\caption{Overview of video editing process with diffusion video encoder}
  %\label{fig:editing}
\end{figure}

%\begin{figure*}[t]
  %\centering
  %\includegraphics[width=\linewidth]{figures/Picture7.pdf}
  %\caption{Comparison to prior video editing methods.}
  %\label{fig:comparison}
%\end{figure*}

\begin{figure*}[t]
  \centering
  \captionsetup[subfigure]{labelformat=empty, position=top}
  \begin{sideways}
    \subfloat[Ours]{\includegraphics[width=0.099\textheight]{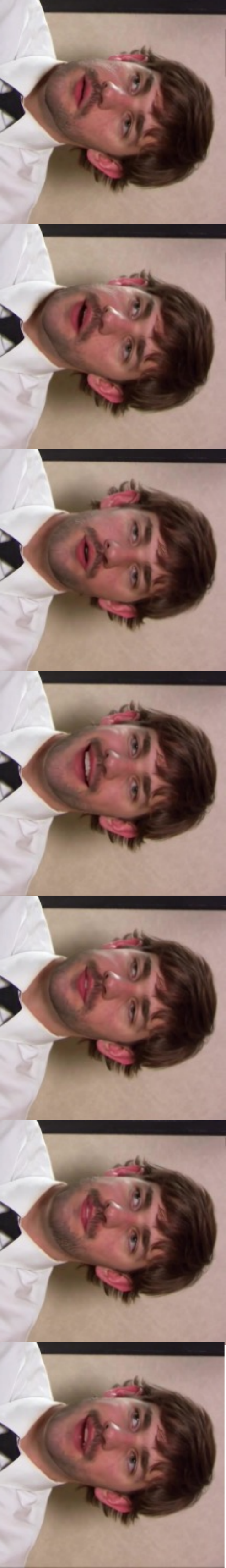}}
    \subfloat[Xu \etal~\cite{eccv22}]{\includegraphics[width=0.099\textheight]{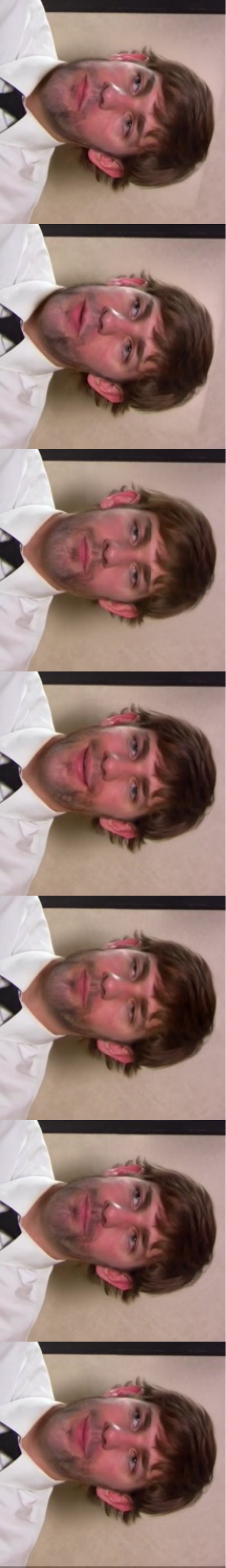}}
    \subfloat[Tzaban \etal~\cite{stit}]{\includegraphics[width=0.099\textheight]{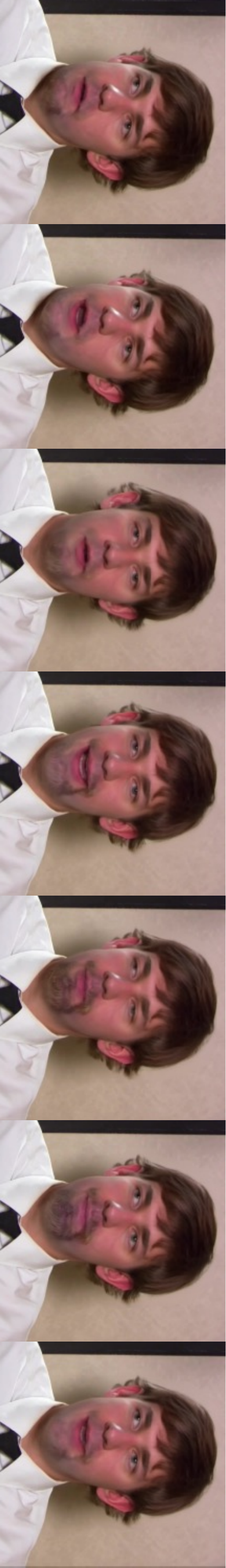}}
    \subfloat[Yao \etal~\cite{latent_transformer}]{\includegraphics[width=0.099\textheight]{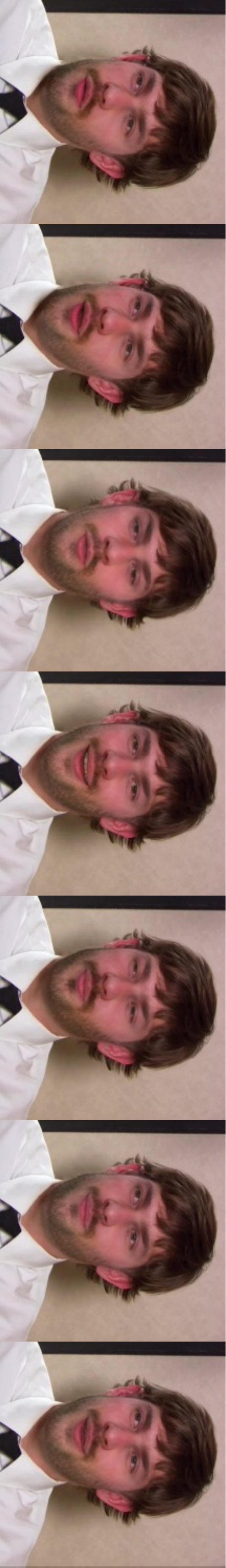}}
    \subfloat[Original]{\includegraphics[width=0.099\textheight]{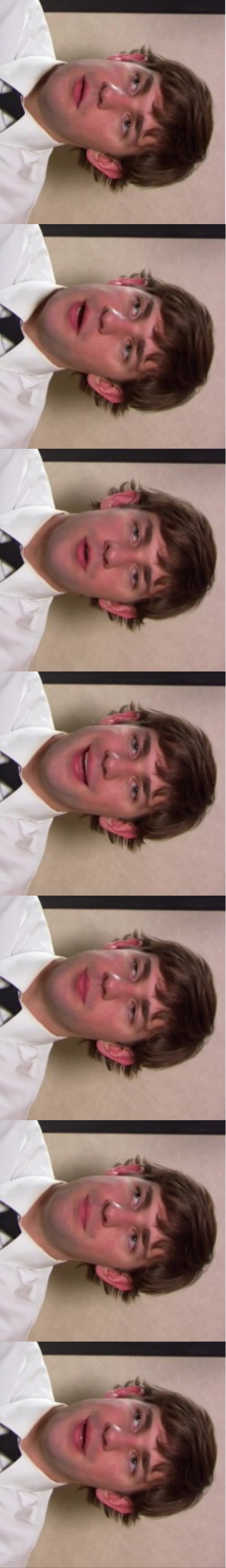}}
  \end{sideways}
  \vspace{-0.05in}
  \caption{Comparison of temporal consistency to the previous video editing methods for `beard'.}
  \label{fig:comparison}
  \vspace{-0.1in}
\end{figure*}

\subsection{Video Editing Framework}
\label{sec:Video Editing Framework}

% Video Editing Framework
We now describe our video editing framework with our diffusion video autoencoders. %The overview can be found in \cref{fig:editing}.
 First, all video frames $\{I_n\}_{n=1}^N$ are aligned and cropped for interesting face regions as in \cite{stit}. The cropped frames $\{x_0^{(n)}\}_{n=1}^N$ are then encoded to latent features using our diffusion video autoencoder. To extract the representative identity feature of the video, we average identity features of each frame as
$$z_\texttt{id,rep}=\frac{1}{N}\sum_{n=1}^{N}(z_\texttt{id}^{(n)}),$$ 
where $z_\texttt{id}^{(n)} = E_\texttt{id}(x_{0}^{(n)})$. Similarly, the per-frame landmark feature is computed as $z_\texttt{lnd}^{(n)}=E_\texttt{lnd}(x_{0}^{( n)})$ to finally obtain per-frame face features $z_\texttt{face}^{(n)}=MLP(z_\texttt{id,rep}, z_\texttt{lnd}^{(n)})$. After that, we compute $x_{T}^{(n)}$ with DDIM forward process conditioning $z_\texttt{face}^{(n)}$.
%Thanks to the disentangled nature of diffusion video autoencoder, aligned and cropped frame $x_{0,1},x_{0,2},\ldots,x_{0,N}$ are encoded to extract the $z_{id,n}, z_{lnd,n},x_{T,n}$ for each frame $x_{0,n}$. Ahead of the editing, the representative identity feature of the video is computed by simply averaging it as $$z_{id,rep}=\frac{1}{N}\sum_{n=1}^{N}(z_{id,n}).$$ 
Thereafter, manipulation is conducted by editing $z_\texttt{id,rep}$ to $z^\texttt{edit}_\texttt{id, rep}$ using a pretrained linear attribute-classifier of the identity space or text-based optimization which we will discuss in more detail later.
After modifying the representative identity feature $z_\texttt{id,rep}^\texttt{edit}$, the edited frame $\hat{x}_{0}^{(n),\texttt{edit}}$ is generated by the conditional reverse process with %$(z_\texttt{id,rep}^\texttt{edit},
$(z_\texttt{face}^{(n), \texttt{edit}},x_{T}^{(n)})$ where $z_\texttt{face}^{(n), \texttt{edit}} = MLP(z_\texttt{id,rep}^\texttt{edit}, z_\texttt{lnd}^{(n)})$. %Afterward, same as in every prior work, the face part of the edited frames are pasted to the corresponding regions of the original frames for clean final product.
Afterward, as with all previous work, the face part of the edited frame is pasted to the corresponding area of the original frame to create a clean final result.
For this process, we segment face regions using a pretrained segmentation network \cite{segmentation-net}.
%with a mask and applied to the original video. The video editing framework through the trained diffusion video autoencoder is summarized in \cref{fig:editing-a}.
% pretrained Classifier based editing
Below, we explore two editing methods  %for identity feature 
for our video editing framework. 

\vspace{-0.1in}
\paragraph{Classifier-based editing} First, as in DiffAE ~\cite{diffae}, we train a linear classifier $C_\texttt{attr}(z_\texttt{id}) = \texttt{sigmoid}(w_\texttt{attr}^\top z_\texttt{id})$ for each attribute $\texttt{attr}$ on CelebA-HQ's~\cite{progressiveGAN} with its attribute annotation in the identity feature space.  % with  annotation and is used for semantic latent space in~\cite{diffae}. 
To change $\texttt{attr}$, we move the identity feature $z_\texttt{id}$ to $\ell_2\texttt{Norm}(z_\texttt{id} + s w_\texttt{attr})$ with a scale hyperparameter $s$.
%using the editing direction that generates a specific attribute.
\vspace{-0.1in}
% CLIP-based editing
\paragraph{CLIP-based editing} Since the pretrained classifier allows editing only for several predefined attributes, we additionally devise the CLIP-guidance identity feature optimization method. Directional CLIP loss \cite{stylegannada} requires two images corresponding to one for neutral text and one for target text, respectively. It implies that we need the synthesized images with our diffusion model, which is costly with full generative process.
%To compute the CLIP loss, we have to synthesize images $\hat{x}_0 = \hat{x}_0(x_T, z^{edit}_{id};\theta)$ from $x_T$ with $z^{edit}_{id}$ that is to be optimized.
%Kim \etal propose an editing method, named DiffusionCLIP, which is to finetune the parameterized model of DDIM or optimizing noise map $x_{t_0}$ at time $t_0$ which is a hyperparameter to minimize CLIP loss. They synthesize an image starting from $x_{t_0}$ with $S$ steps which is $S << t_0$ to compute CLIP loss and also propose another efficient method. % is computed in the image space.% in DDIM,
%However, there is still room for improvement of editing quality due to the absence of meaningful representation of unconditional DDIM. Since it is confirmed that realistic editing is possible using semantic latent in \cite{diffae}, we propose a method of optimizing semantic latent $z_{id}$ through CLIP loss.
Therefore, to reduce the computational cost, we use the drastically reduced number of steps $S ~(\ll T)$ for image sampling.
%schedule. first update the sampling schedule by reducing the entire time step $T$ to the few $n$
In other words, we consider the time steps $t_1,t_2, \ldots, t_S$ where $0 = t_0 < t_1<\ldots<t_S \leq T$ and evenly split $T$. Thereafter, we compute $x_{t_S}$ from the given image $x_0$ that we want to edit (normally chosen as the first frame $x_0^{(1)}$ of the video) through $S$-step of forward process with $z_\texttt{id}=E_\texttt{id}(x_0)$.
Through the sequential reverse steps from $x_{t_S}$, we recover $\hat{x}_{t_s}$ for each time $t_s$ where $s=(S-1), ..., 0$ with the original $z_\texttt{id}$.
Meanwhile,  $x^\texttt{edit}_{t_s}$ are obtained by the single reverse step from $\hat{x}_{t_{s+1}}$ but with variable $z_\texttt{id}^\texttt{opt}$ being optimized initialized to $z_\texttt{id}$ (See \cref{fig:editing}). 
Finally, we  minimize the directional CLIP loss \cite{stylegannada} between intermediate images $\hat{x}_{t_s}$ and $x_{t_s}^\texttt{edit}$, which are still noisy, for all $s$ with a neutral (\eg ``face") and a target text (\eg ``face with eyeglasses").
We choose intermediate images  $x_{t_s}^\texttt{edit}$, $\hat{x}_{t_{s}}$ from $\hat{x}_{t_{s+1}}$ instead of estimated $x_0$ to compute CLIP loss because to
estimate $x_0$ directly from $x_{t_s}$ is incomplete and erroneous for large $t_s$. %Therefore, we conservatively choose to use  $x_{t-1}$ instead of the estimated of $x_0$ while expecting it would be helpful to find the editing direction more stably.
Therefore, we expect that conservatively choosing intermediate images will help to find the edit direction more reliably. We refer the reader to \cref{sec:ablation_noisy_clip} for the ablation study of our noisy CLIP loss.
%In the process of estimating $x_0$ at $x_t$, since the larger $t$ will make a more incomplete result, we think using $x_{t-1}$ would be helpful to find the editing direction more stably than using $x_0$. 
%We demonstrate that CLIP loss can be measured with a noisy image in \B{Sec. XX}.
In addition to the CLIP loss, to preserve the remaining attributes, we also use ID loss (cosine distance between $z_\texttt{id}$ and $z_\texttt{id}^\texttt{opt}$) %as in DiffusionCLIP \cite{diffusionclip}
and  $\ell_1$ loss between face parts of $\hat{x}_{t_s}$ and $x_{t_s}^\texttt{edit}$ for all $s$. After all, the learned editing step $\Delta z_\texttt{id} = z_\texttt{id}^\texttt{opt} - z_\texttt{id}$ is applied to $z_\texttt{id,rep}$.

\begin{table}[t]
\caption{\textbf{Quantitative reconstruction results} on the randomly chosen 20 videos in VoxCeleb1 test set. The reported values are the mean of the averaged per-frame measurements for each video.}
\vspace{-0.2in}
\begin{center}
\setlength{\columnsep}{0.1pt}
\resizebox{\linewidth}{!}{
\begin{tabular}{lcccc}
\toprule
Method & SSIM $\uparrow$
& MS-SSIM $\uparrow$ & LPIPS $\downarrow$ & MSE $\downarrow$ \\
\cline{1-5}

e4e \cite{e4e}   & 0.509 & 0.761  & 0.157  & 0.037 \\
PTI \cite{pti} & 0.765 & 0.939 & 0.063 & 0.007 \\
\cline{1-5}
Ours ($T=20$) & 0.540 & 0.905 & 0.228 & 0.016 \\
Ours ($T=100$) & \textbf{0.922} & \textbf{0.989} & \textbf{0.045} & \textbf{0.002} \\ 
\bottomrule
\end{tabular}
}
\end{center}
\label{table:recon}
\vspace{-0.15in}
\end{table}

\begin{table}[ht]
\caption{\textbf{Quantitative results} to evaluate temporal consistency. 
%\R{TL-ID and TG-ID imply local consistency between adjacent frames and global consistency across all frames, respectively, in terms of identity.} 
Ours show the best global coherency and comparable local consistency to the baselines.}
\vspace{-0.2in}
\begin{center}
\setlength{\columnsep}{0.1pt}
\resizebox{0.65\linewidth}{!}{
\begin{tabular}{lcc}
\toprule
Method & TL-ID & TG-ID  \\
\cline{1-3}
Yao \etal \cite{latent_transformer} & 0.989 & 0.920 \\
Tzaban \etal \cite{stit} & 0.997 & 0.961\\
Xu \etal \cite{eccv22} & 1.002 & 0.983 \\
Ours & 0.995 & 0.996 \\

\bottomrule
\end{tabular}
}
\end{center}
\label{table:consistency}
\vspace{-0.2in}
\end{table}

\section{Experiments}
% Dataset, model, segmentation_net 등

In this section, we present the experimental results to confirm reconstruction performance (\cref{sec:Reconstruction}), temporally consistent editing ability (\cref{sec:TemporalConsistency}), robustness to the unusual samples (\cref{sec:Real video editing}), and disentanglement of the encoded features (\cref{sec:Decomposed features analysis}) with further ablation study (\cref{sec:Ablation study}). 
For this purpose, we train our diffusion video autoencoder on 77294 videos of the VoxCeleb1 dataset~\cite{voxceleb}. %to train the diffusion video autoencoder. 
As preprocessing, frames of the video are aligned and cropped as in \cite{stit} like the FFHQ dataset. Next, they are resized to the size of $256^2$. For the noise estimator $\epsilon_\theta$, the UNet improved by \cite{improved_ddpm} is used, and we train diffusion video autoencoder for 1 million steps with a batch size of 16. Please see more details in \cref{sec:Detailed_Exp_setting}.
%Since there are samples of poor image quality in the data, the diffusion video encoder is trained with a size of $256^2$ images.
%For the remaining part of this section, we verify 
%The remaining sections, it is intended to verify 
%the strength of the diffusion video autoencoder through various experiments, such as video reconstruction (\cref{sec:Reconstruction}), temporally consistent video editing (, feature analysis, ablation study, etc. 

\subsection{Reconstruction}
\label{sec:Reconstruction}
% Table
For video editing, the ability to reconstruct the original video from the encoded one must be preceded. Otherwise, we will lose the original one before we even edit the video. To compare this ability with baselines quantitatively, we measure frequently used metrics for reconstruction - SSIM \cite{ssim}, Multi-scaled (MS) SSIM~\cite{msssim}, LPIPS~\cite{lpips}, and MSE - on randomly selected 20 videos in the test set of VoxCeleb1. As baselines, 
we compared our model with the GAN-based inversion method e4e~\cite{e4e} and PTI~\cite{pti} used by Yao \etal~\cite{latent_transformer} and Tzaban \etal~\cite{stit}, respectively. 
%\R{For PTI, the implementation of Tzaban \etal~\cite{stit} are used with its default hyperparameters.} 
Since StyleGAN-based methods handle the higher resolution images with the size of $1024^2$, we resize the reconstructed results to $256^2$ for comparison. For our method, we vary the number of diffusion steps $T$ to observe computational cost and image quality trade-offs. 
In \cref{table:recon}, our diffusion video autoencoder with $T=100$ shows the best reconstruction ability and still outperforms e4e with only $T=20$. 

%to completely restore the input video as well as the editing method. 
%In particular, since diffusion video autoencoder uses only the average of identity features, assuming that the identity features in all frames will be very  similar, it is necessary to check the possibility of leakage of identity information during editing. Therefore, we also experiment that decoding using the identity feature of each frame for comparison. We evaluate the reconstruction performance in \cref{table:recon} with metrics such as \TODO{~}.

\subsection{Temporal Consistency}
\label{sec:TemporalConsistency}
% Qualitative & Quantitative

In \cref{fig:comparison}, a visual comparison is conducted to evaluate the video editing performance of our diffusion video autoencoder qualitatively. We edit the given video to generate a beard through text-based guidance except for Yao \etal \cite{latent_transformer}, which only allows to edit predefined attributes, by moving to the opposite direction of ``no\_beard". As a result, we demonstrate that only our diffusion video autoencoder successfully produces the consistent result. Specifically, Yao \etal ~\cite{latent_transformer} fail to preserve the original identity due to the limitations of GAN inversion. In the result of Tzaban \etal~\cite{stit}, the shape and the amount of beard constantly changes according to the lip motion. Although Xu \etal \cite{eccv22} show better but not perfect consistency, the motions unintentionally change as a side effect. 
The inconsistency can be observed more clearly between the second and the fifth column except for ours showing a reliable result.

%which can be observed in the second column more clearly. %also varies the degree of beard continually, and as 
%we can see more clearly by comparing the second frame and the fifth frame, all methods except ours are inconsistent. 
%Please note that we edit with CLIP-based latent optimization and the editing direction is determined based on only the first frame in contrast to others that perform the per-frame editing. 
Furthermore, we quantitatively evaluate the temporal consistency of \cref{fig:comparison} %how consistent our editing results are compared
%to the baselines 
in \cref{table:consistency}. Although there is no perfect metric for temporal consistency of videos, 
TL-ID and TG-ID \cite{stit} imply the local and global consistency of identity between adjacent frames and across all frames, respectively, compared to the original. 
These metrics can be interpreted as being %more
consistent as the original is when their values are close to 1. We emphasize that we greatly improve global consistency. TL-ID of Xu \etal \cite{eccv22} is larger than 1 because the motion of the editing results shrinks so that the adjacent frames become closer to each other than the original. Additional quantitative comparisons of temporal consistency, editing quality, and time cost are summarized in \cref{sec:additional_quantitative}.
   % \TODO{Appendix..?}. 

\subsection{Editing Wild Face Videos}
\label{sec:Real video editing}
% edge case

\begin{figure}
  \centering
  \captionsetup[subfigure]{labelformat=empty, position=bottom}
    \subfloat[Input]{\includegraphics[width=0.25\linewidth]{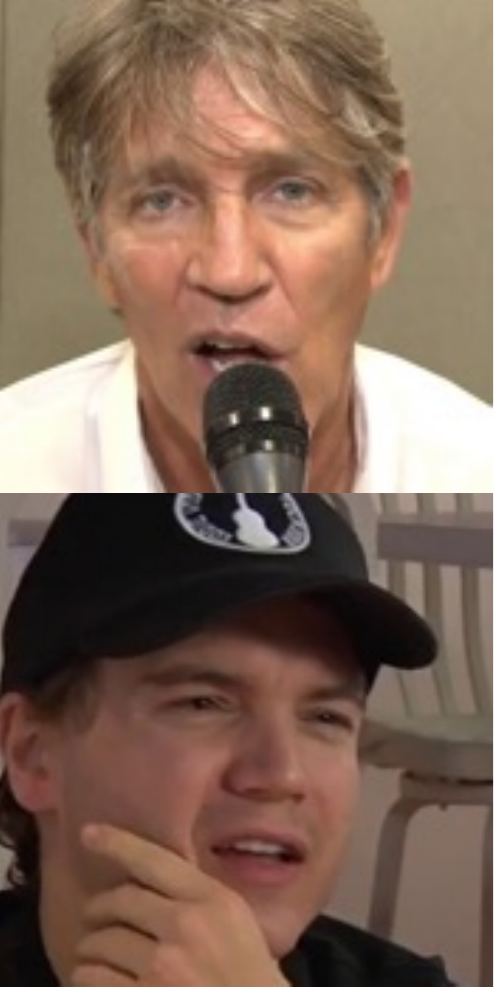}}
    \subfloat[Yao \etal~\cite{latent_transformer}]{\includegraphics[width=0.25\linewidth]{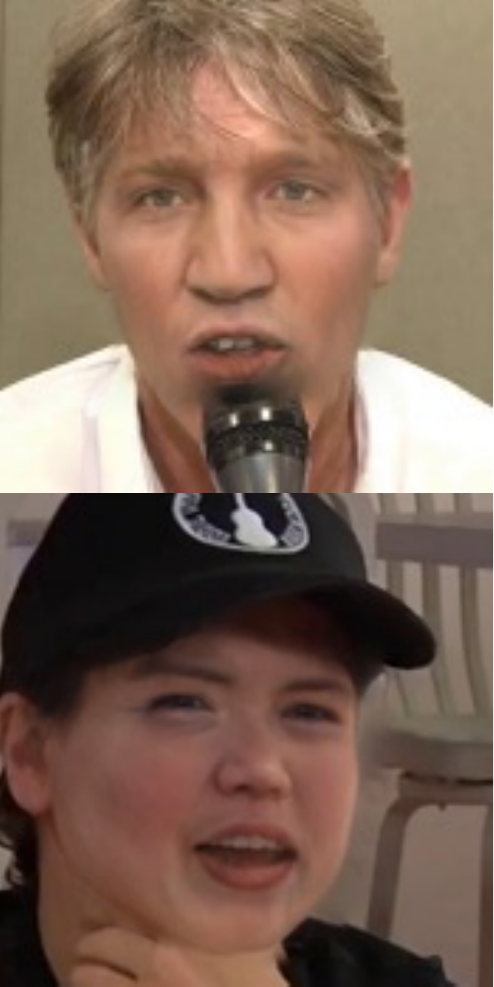}}
    \subfloat[Tzaban \etal~\cite{stit}]{\includegraphics[width=0.25\linewidth]{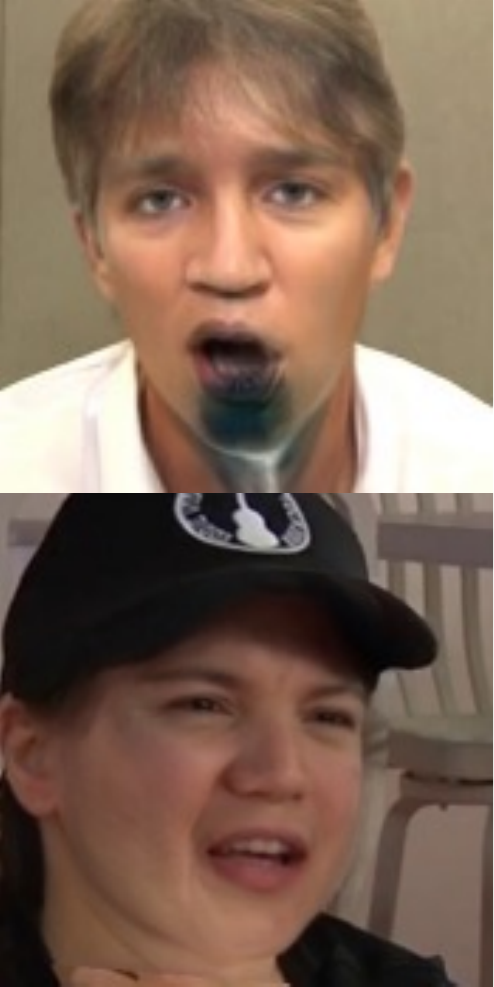}}
    \subfloat[Ours]{\includegraphics[width=0.25\linewidth]{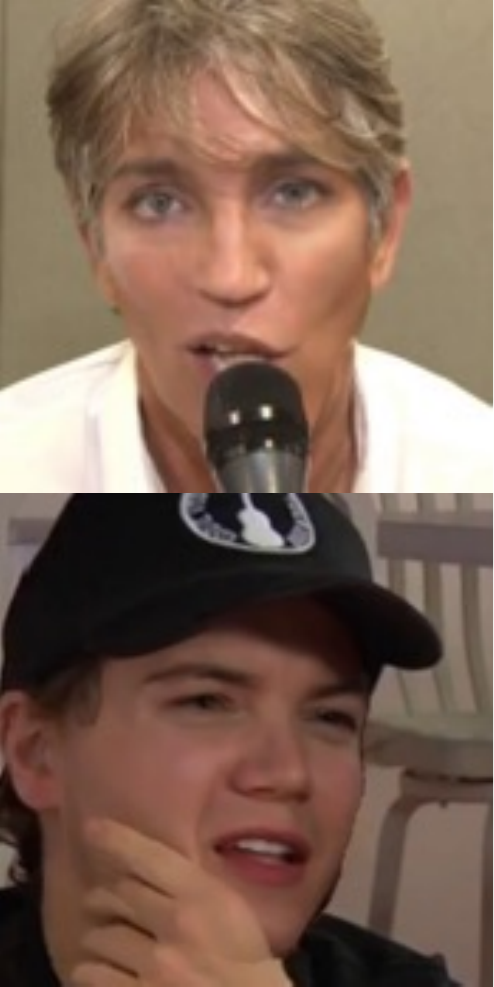}}
    \vspace{-0.05in}
  \caption{Editing wild face videos that GAN-based prior works struggled with. Classifier-based editing is used to make the person "young" (up) or change a "gender" (below).}
  \label{fig:edge-case}
  \vspace{-0.2in}
\end{figure}

Owing to the reconstructability of diffusion models, editing wild videos that are difficult to inversion by GAN-based methods becomes possible. %Therefore, we perform the edge case video editing with \cite{stit} and \cite{latent_transformer} which use GAN-based inversion methods \cite{e4e} and \cite{pti}, respectively. 
As shown in \cref{fig:edge-case}, unlike others, our method robustly reconstructs and edits the given images effectively. % that not only reconstruction but also realistic modification is possible. % young, gender

\subsection{Decomposed Features Analysis}
\label{sec:Decomposed features analysis}
% decomposed feature analysis

\begin{figure}
  \centering
  \includegraphics[width=\linewidth]{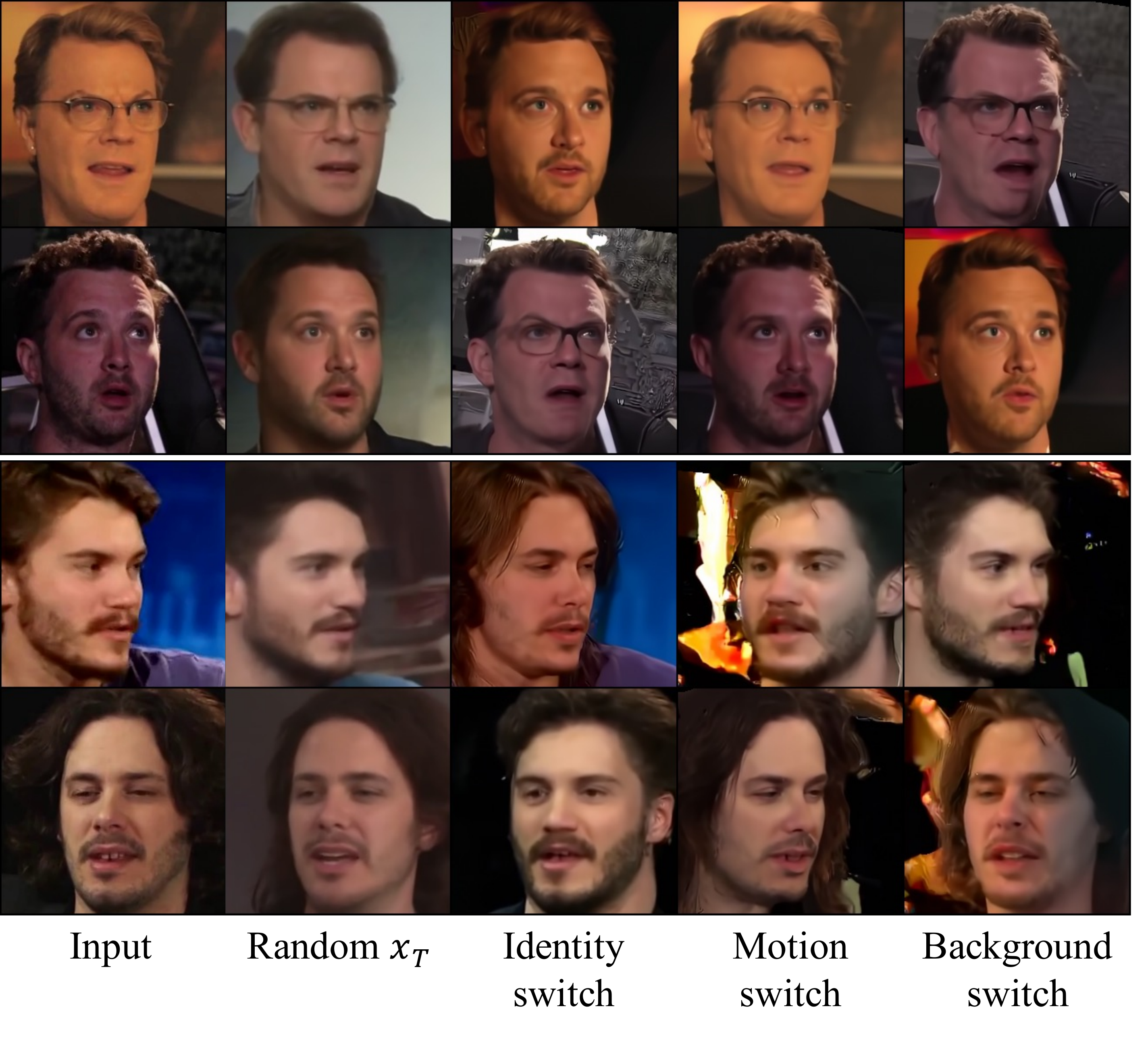}
  \vspace{-0.22in}
  \caption{Demonstration of the disentangled video encoding.}
  \label{fig:decompose}
  \vspace{-0.15in}
\end{figure}

To demonstrate whether the diffusion video autoencoder decomposes the features adequately, % latent space has disentangled property, 
we examine the synthesized images by changing each element of the decomposed features. To this end, we encode the frames of two different videos and then generate samples with a random noise or exchange the respective elements with each other in \cref{fig:decompose}. 
%This result is shown in \cref{fig:decompose}. 
When we decode the semantic latent $z_\texttt{face}$ with a Gaussian noise instead of the original noise latent $x_T$, it has a blurry background that is different from the original one, while identity and head pose are preserved considerably. This result implies that $x_T$ contains only the background information, as we intended. Moreover, the generated images with switched identity, motion, and background feature confirm that the features are properly decomposed and the diffusion video autoencoder can produce a realistic image even with the new combination of the features.
%between the two videos and decode it with a new condition. By reflecting the identity, motion, and background information of different videos well, it is confirmed that the feature is properly decomposed and the diffusion video autoencoder can produce a realistic image even with a modified latent.
However, with extreme changes in the head position, the background occluded by the face is not properly generated because the taken $x_T$ has no information about background in that area (See the last column of the lower example in \cref{fig:decompose}).

%the influence of the original noise latent $x_T$  % where the face was previously located, not the background. 

\subsection{Ablation Study}
\label{sec:Ablation study}
% Encoder 구조 및 xT reg loss 관련
% xT reg loss 
% CLIP loss ablation (x0_esti vs xt) 

\begin{figure}
  \centering
  \includegraphics[width=\linewidth]{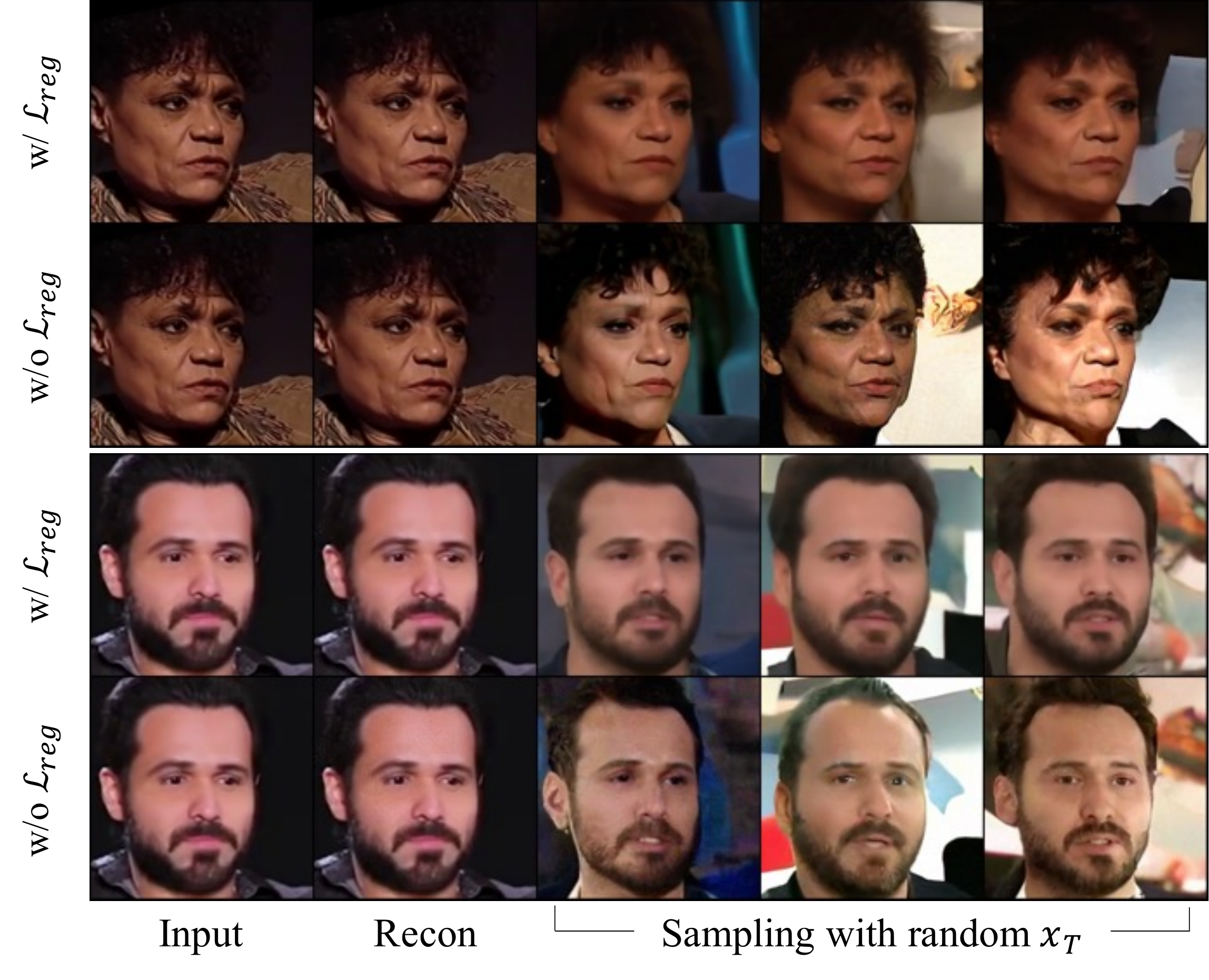}
  \vspace{-0.2in}
  \caption{Ablation of using regularization loss for training. Regularization loss helps the model to contain identity and motion information in $z_\texttt{face}$ as much as possible.}
  \label{fig:ablation-reg}
  \vspace{-0.15in}
\end{figure}

We further conduct an ablation study on our proposed regularization loss $\mathcal{L}_\texttt{reg}$. % and noisy CLIP loss.
First, %to verify the effectiveness of the regularization loss, 
we train the model with the same setting but ablating $\mathcal{L}_\texttt{reg}$. %but only without using the regularization loss.
%Due to the purpose of regularization loss described in \cref{sec:Disentangled Video Encoding}, the results of conditional sampling from random noises using the same $z_\texttt{face}$ should be similar, especially in face part. 
As shown in \cref{fig:ablation-reg}, neither model has a problem with reconstruction. However, without the regularization loss, the identity changes significantly according to the random noise. Thus we can conclude that the regularization loss helps the model to decompose features effectively.

%Secondly, we optimize the latent $z_{id}^{opt}$ by computing the CLIP loss with the estimated $x_0$ at $x_{t_s}^\texttt{edit}$, instead of itself for ablation of noisy CLIP loss. In \TODO{noisy clip figure}, 
%Lastly, since motion and background feature can be stored in both semantic and noise map space, we verify if there is a more suitable place for storing each video feature in the latent space. Note that identity feature must be stored in semantic space for editing. As a result, we confirm that our design choice was best and experimental details are in \TODO{Appendix}

%Finally, the ablation study for the design choice of the diffusion video autoencoder’s latent space. Through experiments, we confirmed which latent space would be the most effective for each of the three features that consist of video, which is summarized in the \TODO{Appendix..?}.

% \section{Limitations and Discussion}
% Image editing의 한계를 뛰어넘을 수 없다는 점, 여자에게서 수염을 생성하는게 쉽지 않다거나 등
% Frozen pre-trained ID encoder에서 오는 한계, Model capacity의 한계로 처리할 수 없는 identity가 있을수도
% Future work (rainable Encoder를 통한 Feature Decompose, Face video에 특정짓지 않고, natural scene video에도 적용해볼 수 있나)

\section{Conclusions}
 
%We have resolved the temporal consistency problem in editing human face videos. %in a different direction from the existing per-frame editing methods. 
%First of all, 
To tackle the temporal consistency problem in editing human face video, we have proposed a novel framework with the newly designed video diffusion autoencoder which encodes the identity, motion, and background information in a disentangled manner and decodes after editing a single identity feature. Through the disentanglement, the most valuable strength of our framework is that we can search the desired editing direction for only one frame and then edit the remaining frames with temporal consistency by moving the representative video identity feature.
%even if the editing direction is determined only in one frame, it can be reflected consistently for all frames. 
Additionally, the wild face video can be reliably edited by the advantage of the diffusion model that is superior in reconstruction to GAN. 
%Therefore, since we propose a novel video editing framework through autoencoding and use the diffusion model for the first time in a video editing task, we can solve both the consistency problem and the Inversion problem in existing video editing methods. 
Finally, %using the fact that the diffusion video autoencoder has a high-level semantic space for identity, 
we have explored to optimize the semantic identity feature with CLIP loss for text-based video editing. %In this process, it is shown that CLIP loss at the noisy image can also be used.
Please refer to the \cref{sec:limitation_discussion} for the limitations and further discussion.

\vspace{-0.2in}
\paragraph{Acknowledgement} This work was supported by Institute of Information \& communications Technology Planning \& Evaluation (IITP) grant (No.2019-0-00075, Artificial Intelligence Graduate School Program(KAIST), No.2021-0-02068, Artificial Intelligence Innovation Hub, No.2022-0-00713, Meta-learning applicable to real-world problems, No.2022-0-00984, Development of Artificial Intelligence Technology for Personalized Plug-and-Play Explanation and Verification of Explanation) and the National Research Foundation of Korea (NRF) grants (No.2018R1A5A1059921) funded by the Korea government (MSIT). This work was also supported by KAIST-NAVER Hypercreative AI Center.

%%%%%%%%% REFERENCES
{\small
\bibliographystyle{ieee_fullname}
\bibliography{egbib}

\begin{thebibliography}{10}\itemsep=-1pt

\bibitem{restyle}
Yuval Alaluf, Or Patashnik, and Daniel Cohen-Or.
\newblock Restyle: A residual-based stylegan encoder via iterative refinement.
\newblock In {\em Proceedings of the IEEE/CVF International Conference on
  Computer Vision}, pages 6711--6720, 2021.

\bibitem{stylegan3video}
Yuval Alaluf, Or Patashnik, Zongze Wu, Asif Zamir, Eli Shechtman, Dani
  Lischinski, and Daniel Cohen-Or.
\newblock Third time's the charm? image and video editing with stylegan3.
\newblock {\em arXiv preprint arXiv:2201.13433}, 2022.

\bibitem{blended}
Omri Avrahami, Dani Lischinski, and Ohad Fried.
\newblock Blended diffusion for text-driven editing of natural images.
\newblock In {\em Proceedings of the IEEE/CVF Conference on Computer Vision and
  Pattern Recognition}, pages 18208--18218, 2022.

\bibitem{landmark}
Cunjian Chen.
\newblock Pytorch face landmark: A fast and accurate facial landmark detector,
  2021.

\bibitem{arcface}
Jiankang Deng, Jia Guo, Niannan Xue, and Stefanos Zafeiriou.
\newblock Arcface: Additive angular margin loss for deep face recognition.
\newblock In {\em Proceedings of the IEEE/CVF conference on computer vision and
  pattern recognition}, pages 4690--4699, 2019.

\bibitem{near_perfect_inv}
Qianli Feng, Viraj Shah, Raghudeep Gadde, Pietro Perona, and Aleix Martinez.
\newblock Near perfect gan inversion.
\newblock {\em arXiv preprint arXiv:2202.11833}, 2022.

\bibitem{stylegannada}
Rinon Gal, Or Patashnik, Haggai Maron, Amit~H Bermano, Gal Chechik, and Daniel
  Cohen-Or.
\newblock Stylegan-nada: Clip-guided domain adaptation of image generators.
\newblock {\em ACM Transactions on Graphics (TOG)}, 41(4):1--13, 2022.

\bibitem{ganspace}
Erik H{\"a}rk{\"o}nen, Aaron Hertzmann, Jaakko Lehtinen, and Sylvain Paris.
\newblock Ganspace: Discovering interpretable gan controls.
\newblock {\em Advances in Neural Information Processing Systems},
  33:9841--9850, 2020.

\bibitem{ddpm}
Jonathan Ho, Ajay Jain, and Pieter Abbeel.
\newblock Denoising diffusion probabilistic models.
\newblock {\em Advances in Neural Information Processing Systems},
  33:6840--6851, 2020.

\bibitem{progressiveGAN}
Tero Karras, Timo Aila, Samuli Laine, and Jaakko Lehtinen.
\newblock Progressive growing of gans for improved quality, stability, and
  variation.
\newblock {\em arXiv preprint arXiv:1710.10196}, 2017.

\bibitem{stylegan2}
Tero Karras, Samuli Laine, Miika Aittala, Janne Hellsten, Jaakko Lehtinen, and
  Timo Aila.
\newblock Analyzing and improving the image quality of stylegan.
\newblock In {\em Proceedings of the IEEE/CVF conference on computer vision and
  pattern recognition}, pages 8110--8119, 2020.

\bibitem{diffusionclip}
Gwanghyun Kim, Taesung Kwon, and Jong~Chul Ye.
\newblock Diffusionclip: Text-guided diffusion models for robust image
  manipulation.
\newblock In {\em Proceedings of the IEEE/CVF Conference on Computer Vision and
  Pattern Recognition (CVPR)}, pages 2426--2435, June 2022.

\bibitem{adam}
Diederik~P. Kingma and Jimmy Ba.
\newblock Adam: A method for stochastic optimization, 2014.

\bibitem{blind}
Wei-Sheng Lai, Jia-Bin Huang, Oliver Wang, Eli Shechtman, Ersin Yumer, and
  Ming-Hsuan Yang.
\newblock Learning blind video temporal consistency.
\newblock In {\em Proceedings of the European conference on computer vision
  (ECCV)}, pages 170--185, 2018.

\bibitem{dpmsolver}
Cheng Lu, Yuhao Zhou, Fan Bao, Jianfei Chen, Chongxuan Li, and Jun Zhu.
\newblock Dpm-solver: A fast ode solver for diffusion probabilistic model
  sampling in around 10 steps.
\newblock {\em arXiv preprint arXiv:2206.00927}, 2022.

\bibitem{dpmsolver++}
Cheng Lu, Yuhao Zhou, Fan Bao, Jianfei Chen, Chongxuan Li, and Jun Zhu.
\newblock Dpm-solver++: Fast solver for guided sampling of diffusion
  probabilistic models.
\newblock {\em arXiv preprint arXiv:2211.01095}, 2022.

\bibitem{sdedit}
Chenlin Meng, Yutong He, Yang Song, Jiaming Song, Jiajun Wu, Jun-Yan Zhu, and
  Stefano Ermon.
\newblock Sdedit: Guided image synthesis and editing with stochastic
  differential equations.
\newblock In {\em International Conference on Learning Representations}, 2021.

\bibitem{voxceleb}
A. Nagrani, J.~S. Chung, and A. Zisserman.
\newblock Voxceleb: a large-scale speaker identification dataset.
\newblock In {\em INTERSPEECH}, 2017.

\bibitem{glide}
Alex Nichol, Prafulla Dhariwal, Aditya Ramesh, Pranav Shyam, Pamela Mishkin,
  Bob McGrew, Ilya Sutskever, and Mark Chen.
\newblock Glide: Towards photorealistic image generation and editing with
  text-guided diffusion models.
\newblock {\em arXiv preprint arXiv:2112.10741}, 2021.

\bibitem{improved_ddpm}
Alexander~Quinn Nichol and Prafulla Dhariwal.
\newblock Improved denoising diffusion probabilistic models.
\newblock In {\em International Conference on Machine Learning}, pages
  8162--8171. PMLR, 2021.

\bibitem{face_decom_map}
Yotam Nitzan, Amit Bermano, Yangyan Li, and Daniel Cohen-Or.
\newblock Face identity disentanglement via latent space mapping.
\newblock {\em arXiv preprint arXiv:2005.07728}, 2020.

\bibitem{styleclip}
Or Patashnik, Zongze Wu, Eli Shechtman, Daniel Cohen-Or, and Dani Lischinski.
\newblock Styleclip: Text-driven manipulation of stylegan imagery.
\newblock In {\em Proceedings of the IEEE/CVF International Conference on
  Computer Vision}, pages 2085--2094, 2021.

\bibitem{diffae}
Konpat Preechakul, Nattanat Chatthee, Suttisak Wizadwongsa, and Supasorn
  Suwajanakorn.
\newblock Diffusion autoencoders: Toward a meaningful and decodable
  representation.
\newblock In {\em Proceedings of the IEEE/CVF Conference on Computer Vision and
  Pattern Recognition}, pages 10619--10629, 2022.

\bibitem{clip}
Alec Radford, Jong~Wook Kim, Chris Hallacy, Aditya Ramesh, Gabriel Goh,
  Sandhini Agarwal, Girish Sastry, Amanda Askell, Pamela Mishkin, Jack Clark,
  et~al.
\newblock Learning transferable visual models from natural language
  supervision.
\newblock In {\em International Conference on Machine Learning}, pages
  8748--8763. PMLR, 2021.

\bibitem{dalle2}
Aditya Ramesh, Prafulla Dhariwal, Alex Nichol, Casey Chu, and Mark Chen.
\newblock Hierarchical text-conditional image generation with clip latents.
\newblock {\em arXiv preprint arXiv:2204.06125}, 2022.

\bibitem{psp}
Elad Richardson, Yuval Alaluf, Or Patashnik, Yotam Nitzan, Yaniv Azar, Stav
  Shapiro, and Daniel Cohen-Or.
\newblock Encoding in style: a stylegan encoder for image-to-image translation.
\newblock In {\em Proceedings of the IEEE/CVF conference on computer vision and
  pattern recognition}, pages 2287--2296, 2021.

\bibitem{pti}
Daniel Roich, Ron Mokady, Amit~H Bermano, and Daniel Cohen-Or.
\newblock Pivotal tuning for latent-based editing of real images.
\newblock {\em ACM Transactions on Graphics (TOG)}, 42(1):1--13, 2022.

\bibitem{stable_diffusion}
Robin Rombach, Andreas Blattmann, Dominik Lorenz, Patrick Esser, and Bj{\"o}rn
  Ommer.
\newblock High-resolution image synthesis with latent diffusion models.
\newblock In {\em Proceedings of the IEEE/CVF Conference on Computer Vision and
  Pattern Recognition}, pages 10684--10695, 2022.

\bibitem{interfacegan}
Yujun Shen, Ceyuan Yang, Xiaoou Tang, and Bolei Zhou.
\newblock Interfacegan: Interpreting the disentangled face representation
  learned by gans.
\newblock {\em IEEE transactions on pattern analysis and machine intelligence},
  2020.

\bibitem{sefa}
Yujun Shen and Bolei Zhou.
\newblock Closed-form factorization of latent semantics in gans.
\newblock In {\em Proceedings of the IEEE/CVF Conference on Computer Vision and
  Pattern Recognition}, pages 1532--1540, 2021.

\bibitem{stylegan_v}
Ivan Skorokhodov, Sergey Tulyakov, and Mohamed Elhoseiny.
\newblock Stylegan-v: A continuous video generator with the price, image
  quality and perks of stylegan2, 2021.

\bibitem{noneq_thermo}
Jascha Sohl-Dickstein, Eric Weiss, Niru Maheswaranathan, and Surya Ganguli.
\newblock Deep unsupervised learning using nonequilibrium thermodynamics.
\newblock In {\em International Conference on Machine Learning}, pages
  2256--2265. PMLR, 2015.

\bibitem{ddim}
Jiaming Song, Chenlin Meng, and Stefano Ermon.
\newblock Denoising diffusion implicit models.
\newblock {\em arXiv preprint arXiv:2010.02502}, 2020.

\bibitem{e4e}
Omer Tov, Yuval Alaluf, Yotam Nitzan, Or Patashnik, and Daniel Cohen-Or.
\newblock Designing an encoder for stylegan image manipulation.
\newblock {\em ACM Transactions on Graphics (TOG)}, 40(4):1--14, 2021.

\bibitem{stit}
Rotem Tzaban, Ron Mokady, Rinon Gal, Amit~H Bermano, and Daniel Cohen-Or.
\newblock Stitch it in time: Gan-based facial editing of real videos.
\newblock {\em arXiv preprint arXiv:2201.08361}, 2022.

\bibitem{face-vid2vid}
Ting-Chun Wang, Arun Mallya, and Ming-Yu Liu.
\newblock One-shot free-view neural talking-head synthesis for video
  conferencing.
\newblock In {\em Proceedings of the IEEE/CVF conference on computer vision and
  pattern recognition}, pages 10039--10049, 2021.

\bibitem{ssim}
Zhou Wang, Alan~C Bovik, Hamid~R Sheikh, and Eero~P Simoncelli.
\newblock Image quality assessment: from error visibility to structural
  similarity.
\newblock {\em IEEE transactions on image processing}, 13(4):600--612, 2004.

\bibitem{msssim}
Zhou Wang, Eero~P Simoncelli, and Alan~C Bovik.
\newblock Multiscale structural similarity for image quality assessment.
\newblock In {\em The Thrity-Seventh Asilomar Conference on Signals, Systems \&
  Computers, 2003}, volume~2, pages 1398--1402. Ieee, 2003.

\bibitem{ode}
Weihao Xia, Yujiu Yang, and Jing-Hao Xue.
\newblock Gan inversion for consistent video interpolation and manipulation.
\newblock {\em arXiv preprint arXiv:2208.11197}, 2022.

\bibitem{eccv22}
Yiran Xu, Badour AlBahar, and Jia-Bin Huang.
\newblock Temporally consistent semantic video editing.
\newblock In {\em European Conference on Computer Vision}, pages 357--374.
  Springer, 2022.

\bibitem{latent_transformer}
Xu Yao, Alasdair Newson, Yann Gousseau, and Pierre Hellier.
\newblock A latent transformer for disentangled face editing in images and
  videos.
\newblock In {\em Proceedings of the IEEE/CVF international conference on
  computer vision}, pages 13789--13798, 2021.

\bibitem{segmentation-net}
Changqian Yu, Changxin Gao, Jingbo Wang, Gang Yu, Chunhua Shen, and Nong Sang.
\newblock Bisenet v2: Bilateral network with guided aggregation for real-time
  semantic segmentation.
\newblock {\em International Journal of Computer Vision}, 129(11):3051--3068,
  2021.

\bibitem{lpips}
Richard Zhang, Phillip Isola, Alexei~A Efros, Eli Shechtman, and Oliver Wang.
\newblock The unreasonable effectiveness of deep features as a perceptual
  metric.
\newblock In {\em Proceedings of the IEEE conference on computer vision and
  pattern recognition}, pages 586--595, 2018.

\end{thebibliography}
}

\clearpage

\appendix
\appendixpage
%\appendix
% \section{Errata}
% \begin{itemize}
%     \item Line 117: Change `These risk' to `This risk'.
%     \item Line 593: Change `$z_\texttt{id} + sw_\texttt{attr}$' to  `$ \ell_2\texttt{Norm}(z_\texttt{id} + sw_{\texttt{attr}})$'
%     \item Line 660: Change `$\hat{x}_{t_{s-1}}$' to `$\hat{x}_{t_s}$'.
%     \item Line 660: Change `from $x_{t_s}$' to `from $\hat{x}_{t_{s+1}}$'.
% \end{itemize}

\section{Detailed Experimental Settings}
\label{sec:Detailed_Exp_setting}

\subsection{Architecture}

We use the model based on the improved version of DDIM \cite{improved_ddpm}. We use linear beta scheduling for $\beta_t$ from $0.0001$ to $0.02$ with $T=1000$. 
This model has UNet structure with the blocks that consist of the residual and the attention blocks. With the number of base channels as 128, the number of channels is multiplied by  $[1,1,2,2,4,4]$ for each block in downsampling layers respectively, and spatial size is down-scaled by half. It is reversed in the upsampling layers. Attention resolution is $[16]$. The dimension of $z_\texttt{face}$ is 512. Time is first embedded into 128 dimensional vector by positional encoding and projected to 512 dimensional vector using a 2-layer MLP with SiLU activation. In each residual block, the time embedding for diffusion modeling and the face feature $z_\texttt{face}$ are first transformed by their corresponding SiLU-Linear layers respectively and these conditions are applied by AdaGN. In more detail, after the input of the residual block is passed to GroupNorm(32)-SiLU-Conv3x3-GroupNorm(32), each channel is scaled and shifted using time embedding. Similarly, after SiLU-Conv3x3-GroupNorm(32), channels are scaled and shifted by the transformed $z_\texttt{face}$. Final block output is obtained after SiLU-Dropout-Conv3x3 following skip connection of the block input. We refer the readers to the implementation code for more details.

\subsection{Training Configuration} 
We optimize the learnable parameters jointly on 77294 videos of VoxCeleb1 dataset \cite{voxceleb}. The videos are aligned and cropped for interesting face regions as in Tzaban \etal~\cite{stit}. We use 4 V100 GPUs and an Adam optimizer \cite{adam} with a learning rate of 1e-4. Total training steps are 1 million and 4 frames per video so a total of 16 frames for 4 videos are taken for a single training step.

\subsection{Manipulation}

\paragraph{Classifier-based editing} The linear classifiers $C_\texttt{attr}$ are trained on CelebA-HQ with attribute annotations in the normalized identity feature space. The classifier is optimized for 10 epochs with the batch size of 32 with a learning rate of 1e-3. Before taken by the classifier, identity features are normalized by the mean and standard deviation of identity features of all samples in VoxCeleb1 test set. Therefore, normalization and denormalization are conducted before and after the identity features are moved by the desired direction $w_\texttt{attr}$ as $\ell_2\texttt{Norm}(\texttt{DeNorm}(\texttt{Norm}(z_\texttt{id}) + sw_\texttt{attr}))$ where $s$ is the hyperparameter for the editing step size, $\texttt{Norm}/\texttt{DeNorm}$ is normalizing and denormalizing function with the statistics of identity features respectively, and $\ell_2\texttt{Norm}$ is the normalization function that makes the $\ell_2$ norm of vectors equal to 1. We use $\ell_2$ normalization because $E_\texttt{id}$ outputs vectors after normalizing their size to 1.

\paragraph{CLIP-based editing} For CLIP-based editing, we use ViT-B/16 among different CLIP architectures. We optimize an identity feature of a single selected frame (usually the first frame of the video) with the Adam optimizer to minimize the CLIP loss. We consider $S=5$ for the number of intermediate latent states. After conduction optimization, the learned editing direction $\Delta z_\texttt{id}$ multiplied by the editing step size is added to the representative identity feature of the video $z_\texttt{id,rep}$. The final edited feature is obtained by applying $\ell_2\texttt{Norm}$. The search spaces of remaining hyperparameters are provided in \cref{table:hyperparam}.

\begin{table}[htb]
\caption{Hyperparameter search space}
\vspace{-0.2in}
\begin{center}
\setlength{\columnsep}{0.1pt}
%\resizebox{\linewidth}{!}{
\begin{tabular}{lc}
\toprule
Parameter & Search space   \\
\cline{1-2}
Learning rate & [2e-3, 4e-3, 6e-3] \\
Weight of CLIP loss & [3] \\
Weight of ID loss & [1, 3, 5] \\
Weight of $\ell_1$ loss & [1, 3, 5] \\
Editing step size & [0.1, 0.5, 1.0, 1.5, 2.0, 2.5] \\
Optimization steps & [2000] \\

\bottomrule
\end{tabular}
%}
\end{center}
\label{table:hyperparam}
\vspace{-0.2in}
\end{table}

%We use adam optimizer to  and the learning rate is tuned among [2e-3, 4e-3, 8e-3]
%\B{몇 iteration 만큼 학습시켰는지, optimizer, lr, hyperparameter space 등}

\section{Ablation of Noisy CLIP Loss}
\label{sec:ablation_noisy_clip}

\begin{figure}
  \centering
  \includegraphics[width=0.8\linewidth]{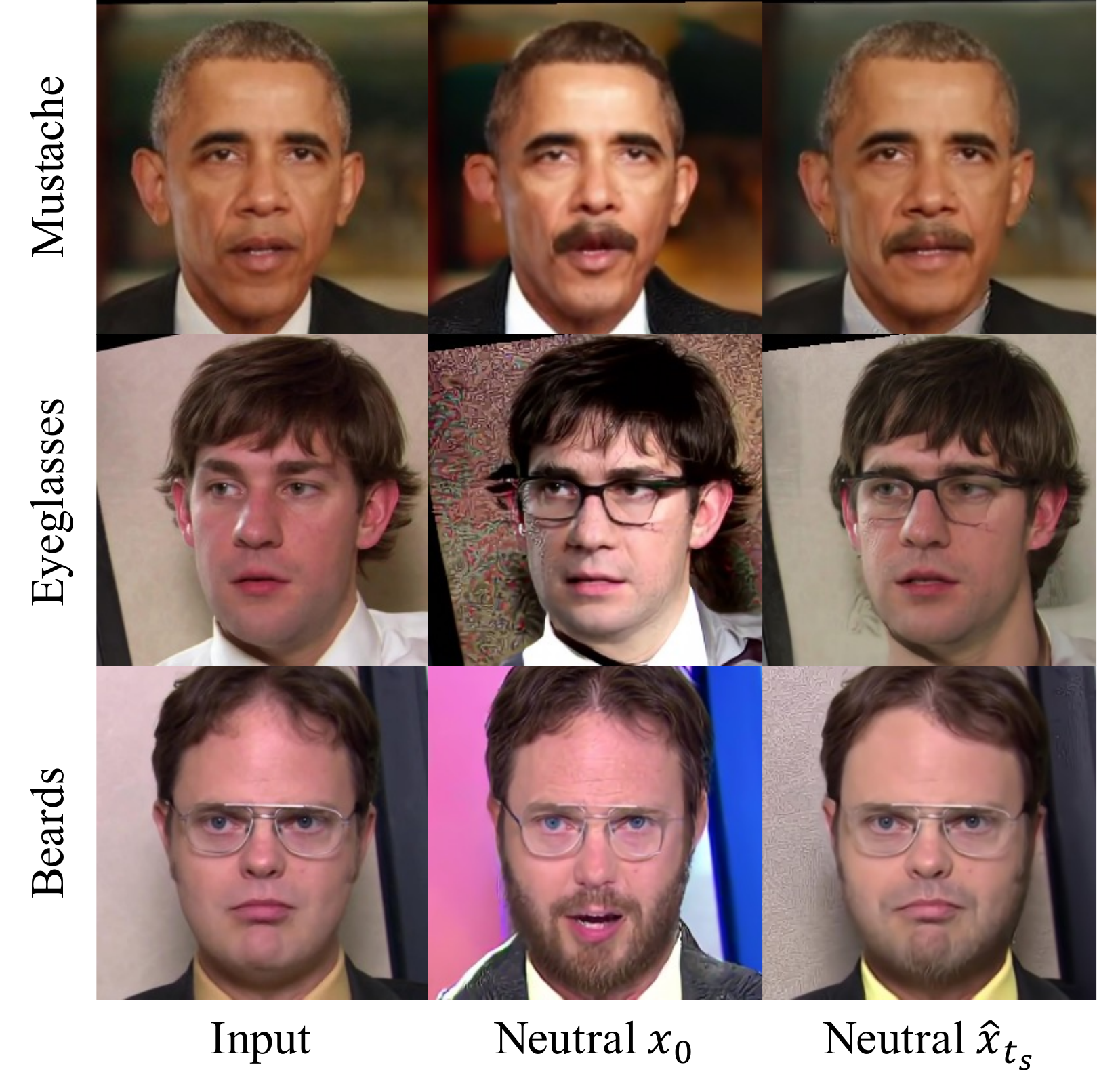}
  \vspace{-0.1in}
  \caption{Ablation of using clean image $x_0$ or intermediate noisy image $\hat{x}_{t_s}$ for a neutral image in directional CLIP loss. As the target image, the former uses estimated $x_0$ at the intermediate time step like \cite{diffusionclip} and the latter uses $x^\texttt{edit}_{t_s}$.}
  \label{fig:ablation-clip}
  \vspace{-0.1in}
\end{figure}

In this section, we conduct an ablation study of our noisy CLIP loss introduced in   \cref{sec:Video Editing Framework}. We compare our CLIP-based editing method using noisy CLIP loss (See \cref{fig:editing}) with the way that uses estimated $x_0$ conditioned by $z_\texttt{id}^\texttt{opt}$ as the target image and the original $x_0$ as the neutral image for each time step $t_s$ to compute the directional CLIP loss, which is similar to the method suggested by Kim \etal \cite{diffusionclip}.  The results are presented in \cref{fig:ablation-clip}. For a fair comparison, we use the same weights for $\ell_1$ loss and ID loss.

%that uses the intermediate outputs $\hat{x}_{t_{s}}$ and $x_{t_s}^\texttt{edit}$ for the neutral and target image respectively for directional CLIP loss (See \cref{fig:ablation-clip}) with 

%between using the clean image $x_0$ and using intermediate result $\hat{x}_{t_{s}}$ for neutral image is summarized in \cref{fig:ablation-clip}. Since the difference between target image $x_{t_{s}}^\texttt{edit}$ and the neutral image $x_0$ in a large time step, optimization becomes unstable.

To help readers understand, we first briefly explain the directional CLIP loss \cite{stylegannada} and DiffusionCLIP \cite{diffusionclip} before we address the ablation results. The directional CLIP loss compares the direction from neutral image embedding to target image embedding with the direction from neutral text embedding to target text embedding in the CLIP space to edit the target image to match the target text. 
%To edit face videos with text prompt, we suggest a noisy CLIP loss that uses intermediate results of generating process for computing the directional CLIP loss~\cite{stylegannada}. 
Kim \etal \cite{diffusionclip} propose an image manipulation method, named DiffusionCLIP, that optimizes the unconditional diffusion model $\epsilon_\theta$ with the directional CLIP loss \cite{stylegannada}.
To preserve the original images to some extent, Kim \etal \cite{diffusionclip} consider the latent states of the original images in not the whole but just a partial range such as $[0, T/2]$ obtained by sparsely passing through the range with the deterministic forward process of DDIM. 
In the GPU-efficient version of DiffusionCLIP, they take an estimated $x_0$ from the latent states in considered intermediate time steps as the target images and compare them with the clean original image $x_0$ as the neutral image. 
%Although Kim \etal~\cite{diffusionclip} propose a GPU-efficient fine-tuning method that uses estimated $x_0$ at the intermediate steps as the target images, they choose the clean original image $x_0$ as the neutral image. 

Going back to the ablation study, unlike Kim \etal \cite{diffusionclip},  %which starts sampling from the middle time step (\eg $T/2$)
we consider the entire range $[0, T]$ to %allow more flexible changes
start conditional sampling from the noise that only has background information and split the range into total $S$ steps (\eg $S=5$) for computational efficiency.
In this case, applying CLIP loss between the original image (neutral) and the estimated $x_0$ (target) makes identity to be altered as in the second column of \cref{fig:ablation-clip} because the difference between the estimated $x_0$ at time $t$ and the clean image $x_0$ becomes larger as $t$ goes larger.
To overcome this phenomenon, we apply CLIP loss between the intermediate outputs $\hat{x}_{t_s}$ (conditioned on original $z_\texttt{id}$ for the neutral images) and $x_{t_{s}}^\texttt{edit}$ (conditioned on trainable $z_\texttt{id}^\texttt{opt}$ for the target images). In the last column of \cref{fig:ablation-clip}, the original identity is well preserved with the desired features edited properly. From these results, we conclude that applying the CLIP loss between the images on the same level of uncertainty as our method leads to relatively stable editing results.

%at each time step $t_s$ as a neutral image %instead of the original $x_0$ 
%and use $x_{t_{s}}^\texttt{edit}$ sampled with a trainable $z_\texttt{id}^\texttt{opt}$ as a target image. 

\section{Additional Quantitative Results}
\label{sec:additional_quantitative}

\subsection{User Study}
% editing quality와 temporal consistency를 위해 user study 진행
% User study exp setting 설명
Current video editing tasks still lack metrics to measure how well the video is edited or whether the edited video is temporally consistent. Although quantitative results have already been reported with metrics used by the prior work~\cite{stit}, we further conduct the user study for sufficient evaluation of editing quality and temporal consistency.
% editing quality를 비교하기 위해서 적절한 metric이 없기 때문에 
%Current video editing tasks still lack perfect metrics. % to measure editing quality and temporal consistency. 
%To measure FVD, we need a lot of generated samples to obtain meaningful statistics, which is infeasible under the time constraint.
% Since our method basically performs editing per frame, it inherits the characteristics and advantages of the latest diffusion models that perform editing at the image level. While we also verified that ours show better performance in terms of consistency with FVD score of 289.36 vs baseline~\cite{stit} 346.48, in order to more reliably evaluate the naturalness and consistency at the video level, we conducted a human evaluation. 
Since we use both classifier-based and CLIP-based editing methods, we choose Tzaban \etal~\cite{stit} which allows both predefined attributes and CLIP-based editing as a baseline. For a fair comparison, the hyperparameters of baseline, $\alpha$ and $\beta$, are carefully determined among the $4, 8, ..., 24$, and $0.1, 0.2, 0.3$ respectively. 

52 volunteers were asked to select the superior result between the edited output of prior work~\cite{stit} (GAN-based) and ours (with $T=100$ for fairness in time-cost) on 24 videos. 
%(refer to \href{https://drive.google.com/drive/folders/1wXVnCk2tA77t9SjSpb2JwQslYvqiXyZq?usp=sharing}{Google Drive} for the used videos). 
The evaluation covers two aspects%, editing quality and temporal consistency
; 1) \textit{quality}: the given target attribute should be properly reflected in the video, and 2) \textit{consistency}: consecutive frames continue naturally after editing.
%with a total of 48 questions
%Instead, we conduct human evaluation that \B{XX} volunteers select the better one between the edited results of GAN-based method STIT \cite{stit} and ours on 24 videos for these two aspects (the total of 48 questions).
%To obtain the examples, we carefully tune the hyperparameters for both methods for each video and a target attribute. 
% FVD 는 비디오 샘플 수 많아야 하는데 시간적인 이유 (많은 샘플 만드는데 시간 오래걸리는 자세한 이유 필요?) 로 측정 불가했음
% latent transformer에서 진행한 실험은
% editing quality와 temporal consistency관련 더 적절한 metric이 없어 human eval 을 진행하였음
%We edit 24 videos each by our method and baseline for various target attributes. 
%Due to the time constraint, we set $T=100$ for our method.
%In the user study, volunteers select which video is better in terms of editing quality and temporal consistency. 
%24 samples (48 questions) are evaluated by ?? volunteers in total.

In \cref{table:user_study}, 61.9\% and 66.3\% of the volunteers favor our method in terms of editing quality and temporal consistency, respectively. %For a fair comparison, we match the inference time costs between ours and the baseline; we reduce iterations in our method.
% Moreover, ??\% of the volunteers judge that our method is more temporally consistent when considering only the target attributes such as beard and eyeglasses, which are especially difficult to maintain temporal consistency with GAN-based methods. 
%When we consider only consistency fragile attributes (\eg beard, eyeglass), 72.3\% of the volunteers determine our method is more temporally consistent.
In particular, certain attributes (such as a beard or eyeglasses; called `fragile' in the \cref{table:user_study}) exhibit a noticeable lack of temporal consistency when edited using the baseline method \cite{stit}.
%Especially, there are some attributes (\eg beard, eyeglasses, etc.;called fragile in the table below) that the edited results by the baseline show temporal inconsistency particularly (see \href{}{Google Drive}). 
When only considering these cases, 72.3\% of the volunteers picked ours for better consistency.

\begin{table}[h]
\vspace{-0.1in}
\caption{Results of a user study}
\vspace{-0.2in}
\begin{center}
%\scriptsize
\setlength{\columnsep}{0.1pt}
\resizebox{0.8\linewidth}{!}{
\begin{tabular}{lccc}
\toprule
& quality &\multicolumn{2}{c}{consistency} \\ %& \multirow{2}{*}{FVD}\\
\cline{2-4} 
Method & all & all & fragile \\
\cline{1-4}
Tzaban \etal~\cite{stit} & 38.1 & 33.7 & 27.7 \\
Ours & \textbf{61.9} & \textbf{66.3} & \textbf{72.3} \\ 
\bottomrule
\end{tabular}
}
\end{center}
\label{table:user_study}
\vspace{-0.1in}
\end{table}

\subsection{Disentangled Editing}
% irrelevant attribute 수정되는지 확인 실험
Although we verified that ours show better performance in terms of temporal consistency in \cref{table:consistency}, this metric cannot detect the identity-irrelevant attribute. Thus, we measured non-target attributes preservation as target attribute change in the way used by Yao \etal~\cite{latent_transformer}. 

\begin{figure}[h]
\centering
\includegraphics[width=0.98\linewidth]{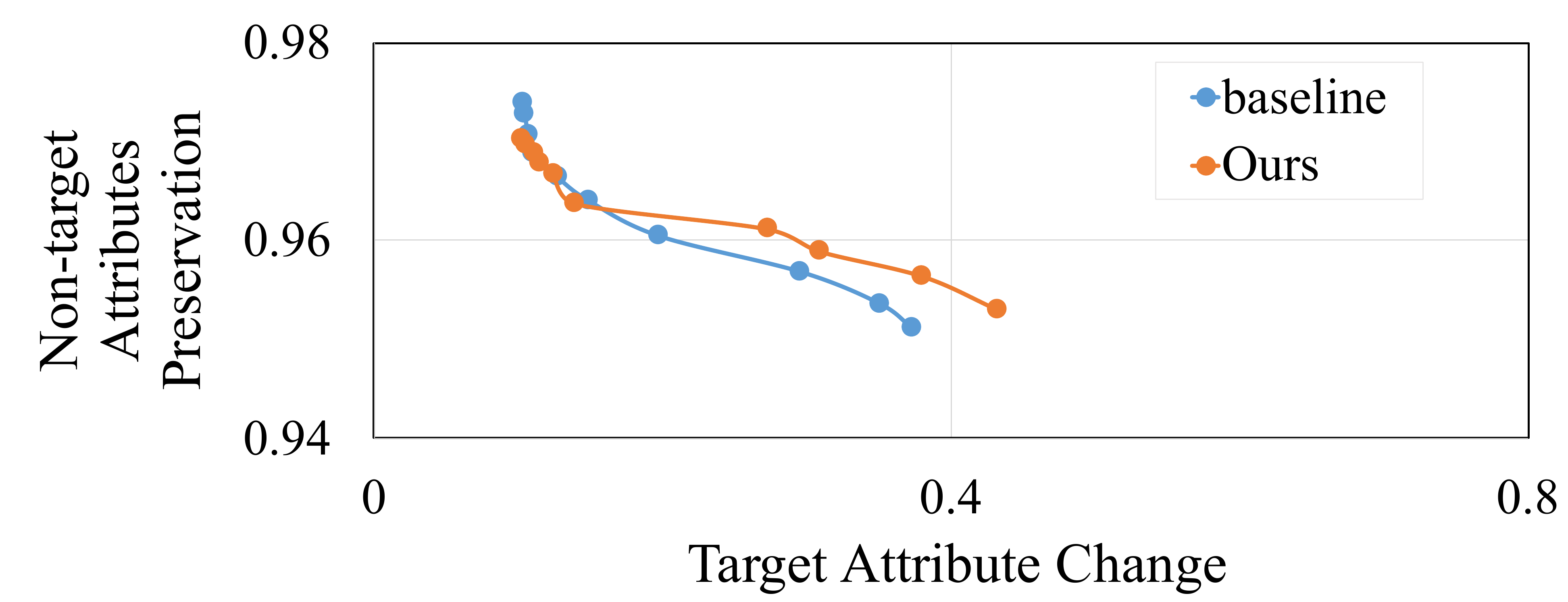}
\vspace{-0.1in}
\caption{Attribute preservation vs. target attribute change (a higher curve is better).}
\label{fig:disentangled-exp}
\vspace{-0.1in}
\end{figure}

Here, we use videos and the corresponding target attributes prepared for the user study.  
Since we measured 24 edited videos with various target attributes, unlike Yao \etal~\cite{latent_transformer} used 1K images for each target attribute, we averaged a non-target attributes preservation rate as corresponding target attribute change.
The \cref{fig:disentangled-exp} shows that our method is slightly better at preserving non-target attributes compared to the baseline~\cite{stit}.

\subsection{Inference Time Comparison}

% todo) time cost
% 1frame을 처리하는데 걸리는 시간 평균 Table로 정리, STIT이 왜 오래걸리는지 + ODE 적용 부분
% google drive에 업로드
While GAN-based methods require only a single pass to synthesize a new edited frame, they internally run PTI~\cite{pti} for reconstruction, which takes more than a few seconds per frame. As shown in the \cref{table:time_cost} where we compare the elapsed time with a single RTX 3090 to edit per frame via ours and %the representative baseline, 
Tzaban \etal \cite{stit}, we are rather slow with full iterations of $T$ but slightly faster at $T=100$ with sufficiently reasonable performance. %(refer to  \href{https://drive.google.com/drive/folders/137rapMeNuv89EkJh_yn7MP_I4h7wZyRC?usp=sharing}{Google Drive}). % showing that the performance of $T=100$ is also sufficiently reasonable. 
Moreover, 
%If the time constraint is our major concern, 
deterministic and faster ODE samplers~\cite{dpmsolver,dpmsolver++} can also be utilized to reduce time further to 2.9s (the 3rd order of DPM-solver with 15 steps) with comparable quality (refer to \href{https://diff-video-ae.github.io}{project page}).

%in the table below. Although, GAN-based methods require only a single pass to synthesize a new edited frame, %they have to go through the
%PTI~\cite{pti} process has to precede for reconstruction, which takes a long time. %existing methods that use GANs, as well as baseline, 
%that they take a long time. 
%On the other hand, our method with $T=100$ can get sufficiently reasonable results and also faster than the baseline. 
% 확인되면
%\B{Deterministic and faster ODE samplers can also be utilized to reduce the time further to XX when applying 3rd order of DPM-solver with 15 steps.}
%%%%%%%%%%
%To demonstrate this qualitatively, we uploaded a video of the results according to $T$ on \href{https://drive.google.com/drive/folders/1if93J3cGhNTb6JGHfZIFD-lpD9o4ZntD?usp=sharing}{Google Drive}.

\begin{table}[h]
\vspace{-0.05in}
\caption{Inference time comparison}
\vspace{-0.2in}
\begin{center}
%\scriptsize
\setlength{\columnsep}{0.1pt}
\resizebox{0.9\linewidth}{!}{
\begin{tabular}{lccc}
\toprule
& \multicolumn{2}{c}{Ours} & \multirow{2}{*}{Tzaban \etal~\cite{stit}} \\
& $T=1000$ & $T=100$ &  \\
\cline{1-4}
Classifier & 60.9s & 5.8s & 12.7s \\
%\cline{1-4}
CLIP & 62.4s & 7.3s & 12.0s\\ 
+ sampler & \multicolumn{2}{c}{\textbf{2.9s}} & \\ 
\bottomrule
\end{tabular}
}
\end{center}
\label{table:time_cost}
\vspace{-0.25in}
\end{table}

\section{Limitations and Further Discussion}
\label{sec:limitation_discussion}

% invariant identity feature를 수정하다보니 motion 불가능
% disentanglement 를 위해 pretrained the ID encoder and the landmark encoder를 사용했기 때문에 이 ID encoder가 표현할 수 있는 것 이외의 attribute에 대해서는 editing 할 수 없다는 단점이 있다. 
%또한, GAN based methods에서도 나타나는 문제인데 여전히 data의 bias에 영향을 받아서 gender bias 가 나타나기도 한다 (여자 얼굴에 수염은 잘 안만들어지는 등). 단순한 linear classifier로는 entangled attribute features 에 대한 manipulation이 동시에 나타나기도 한다. 

%The main limitations of our method come from exploiting the pretrained networks such as an ID encoder (ArcFace) and a landmark encoder. 1) Using these networks limits the domain to face video as other previous works including face editing, face swapping, and face reenactment. 2) Although the ID encoder enables us to decompose time-dependent and -invariant identity features, we can not fully disentangle them. \textcolor{blue}{TODO: why we fail to disentangle.}might be 3)?
% Limitations
% First, our editing method, like other baselines, is limited to modifying attributes within the face. 
% Although the pretrained ID encoder enables us to decompose time-dependent and -invariant identity features, it sometimes limits the performance: i) some edits do not work well with the ID encoder. For example, facial expressions are more about motion than identity. We conjecture that this is the reason for the failure case that the reviewer \RC ~pointed out. ii) the latent space of the pre-trained ID encoder is not disentangled as much as the style space of StyleGAN. For example, we observe a gender bias when attempting to apply a `beard' to Emma Watson (see \href{https://drive.google.com/drive/u/5/folders/1krtAkUGJER9oHoV1IGzbleeCS4quBN7z}{Google Drive}). 
The main limitations of our method come from exploiting the pretrained networks such as an identity encoder (ArcFace) and a landmark encoder: 1) Using these networks limits the domain to face video as other previous works including face editing, face swapping, and face reenactment. 2) Our method is difficult to edit poses or facial expressions that can not be fully captured by the identity encoder. We conjecture that this is the reason why the eyebrows are unnatural in the last row of \cref{fig:add-clip-vox}. 
%the reviewer \RC ~pointed out. 
3) Since the identity encoder is trained for face recognition tasks, the latent space may lack disentanglement for editing. For example, we observed a gender bias when attempting to apply a `beard' to a woman.
%Emma Watson (see \href{https://drive.google.com/drive/folders/1krtAkUGJER9oHoV1IGzbleeCS4quBN7z?usp=sharing}{Google Drive}). 
As a future direction, this weakness could be resolved by finding out the disentangled space analogous to the style space of StyleGAN or training a module to discover disentangled editing directions. 

In addition, a higher resolution video is possible as a future direction. We apply our method to $256^2$ resolution videos for the following reasons: 1) The implementation of diffusion autoencoders \cite{diffae} on which we are based is for $256^2$ images. 2) The dataset used, VoxCeleb1, includes many low-resolution videos, which we have resized to a suitable and balanced size of $256^2$. For a higher resolution, our method can be seamlessly applied by exploiting a diffusion upsampler module as DALLE-2 \cite{dalle2} or latent diffusion model architecture as Stable Diffusion~\cite{stable_diffusion} with conditioning our semantic representation. 

%we can expect that there may be attributes where it is unable to express or edit with the pretrained ID encoder which might degrade the editability. Representatively, facial expressions are more about motion than identity. We conjecture that this is the reason why the failure case that the reviewer \RC ~pointed out (the last row in Fig. 12) shows an unnatural result. Furthermore, the latent space of the pre-trained ID encoder that our methid is using is not disentangled as much as the style space of StyleGAN, resulting in some entanglement of correlated attributes.
%Consequently, it is observed that there is a gender bias when attempting to apply a `beard' to a video of Emma Watson, as the video is transformed into that of a male. \B{However, ours still demonstrates incomparable temporal consistency to the baseline.} The edited video can be found in \href{https://drive.google.com/drive/u/5/folders/1krtAkUGJER9oHoV1IGzbleeCS4quBN7z}{Google Drive}. As a future direction, this weakness could be resolved by finding out the disentangled space analogous to the style space of StyleGAN or develop a method of distentangled editing direction search.

\section{Comparison of Temporal Consistency}
We upload the video file of \cref{fig:comparison} to the \href{https://diff-video-ae.github.io}{project page}. In the video, the result of Yao \etal \cite{latent_transformer} shows an altered identity, the result of Tzaban \etal \cite{stit} shows temporal inconsistency that beards fade away as the mouth opens, and the result of Xu \etal \cite{eccv22} shows unnatural movements with the mouth not opening as much as the original and inconsistency of the beard. On the other hand, ours demonstrates much improvement in terms of the temporal consistency and identity preservation.

\section{Additional Editing Results}
\begin{figure*}
  \centering
  \includegraphics[width=0.98\linewidth]{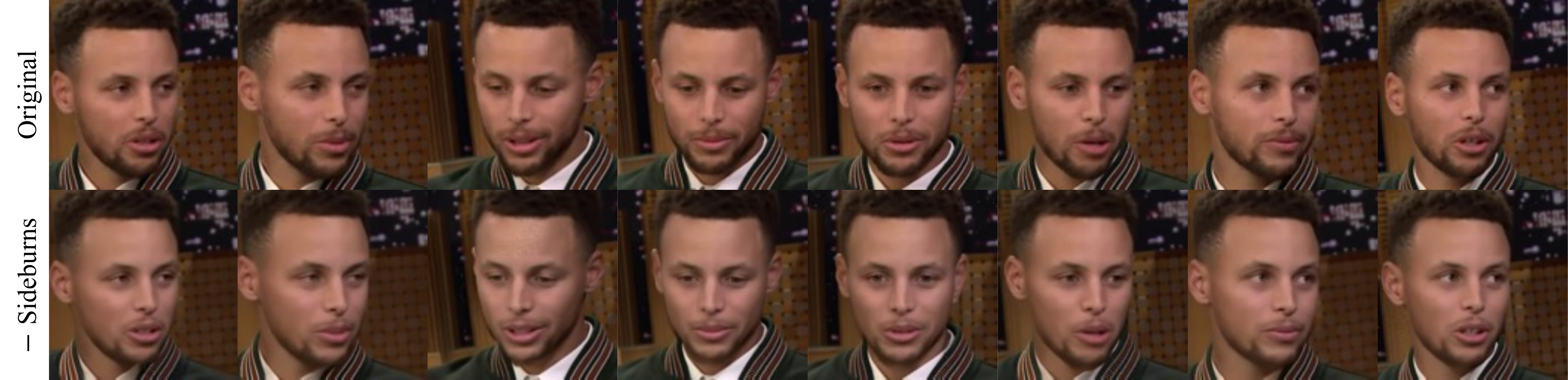}
  \includegraphics[width=0.98\linewidth]{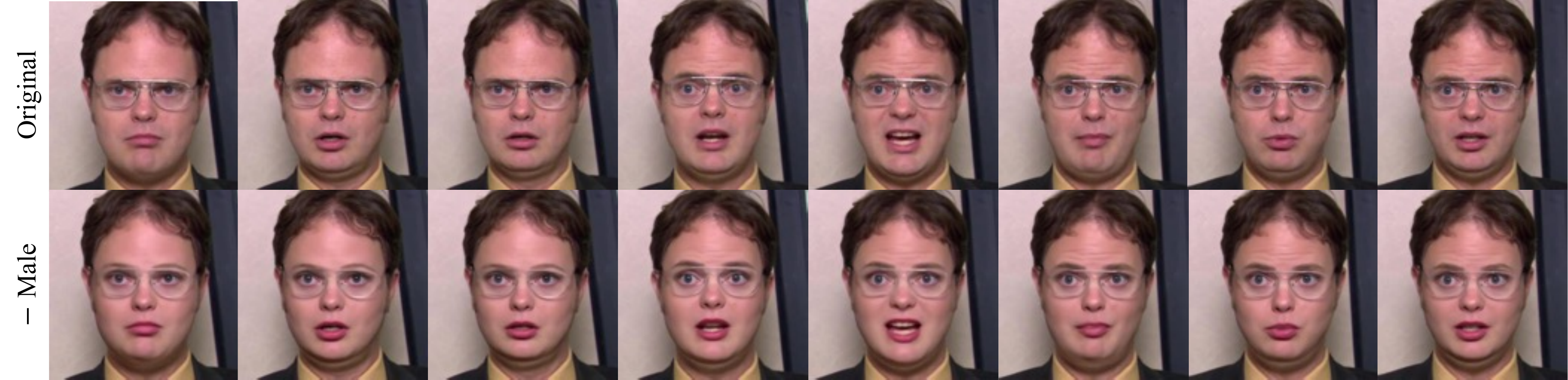}
  \includegraphics[width=0.98\linewidth]{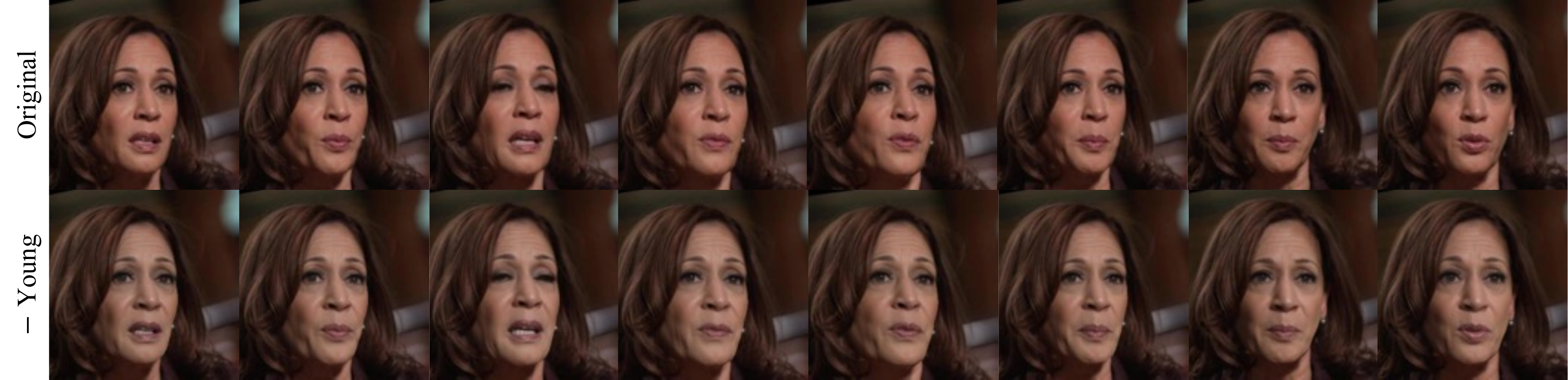}
  \includegraphics[width=0.98\linewidth]{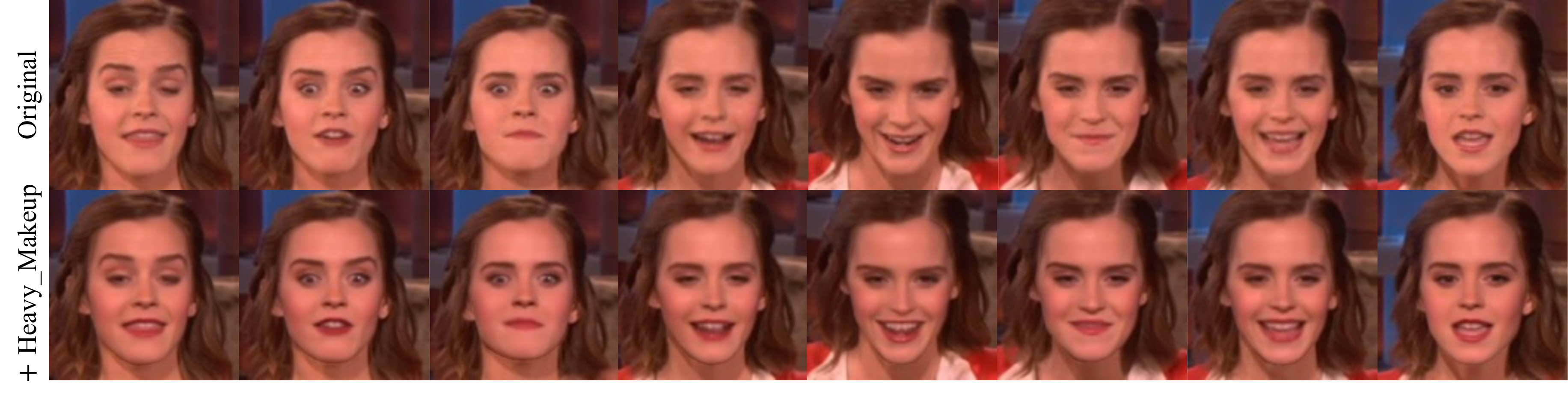}
  \caption{Classifier-based video editing on the other videos not in VoxCeleb1.}
  \label{fig:add-classifier-stit}
\end{figure*}

\begin{figure*}
  \centering
  \includegraphics[width=0.98\linewidth]{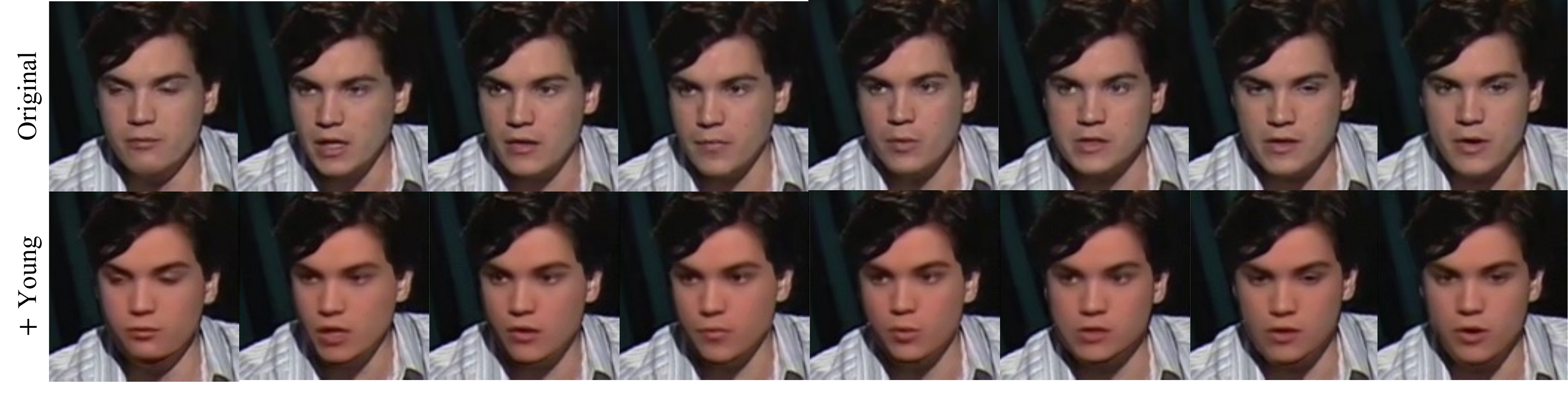}
  \includegraphics[width=0.98\linewidth]{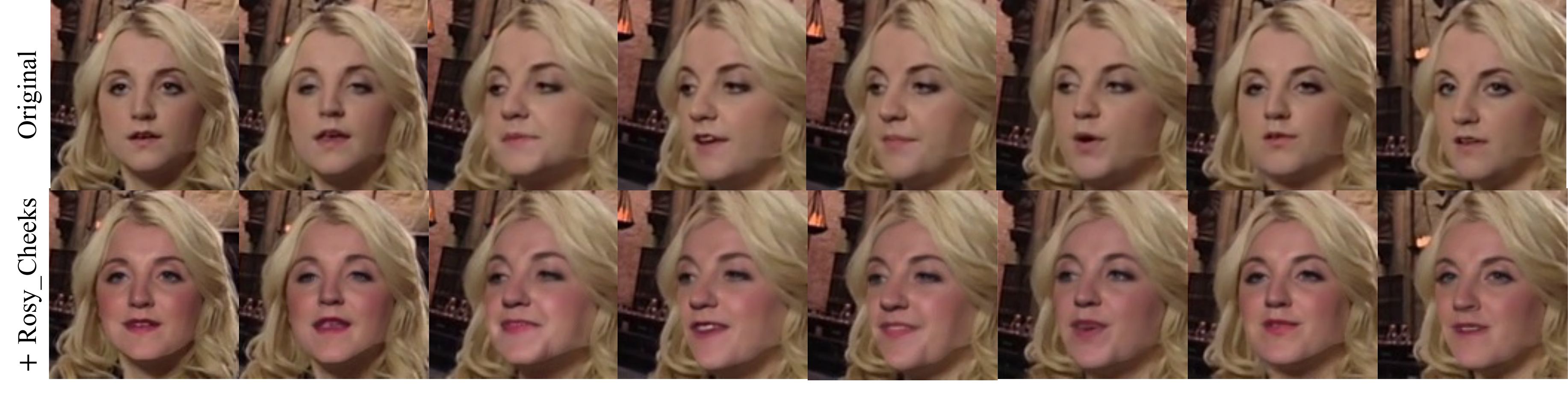}
  \includegraphics[width=0.98\linewidth]{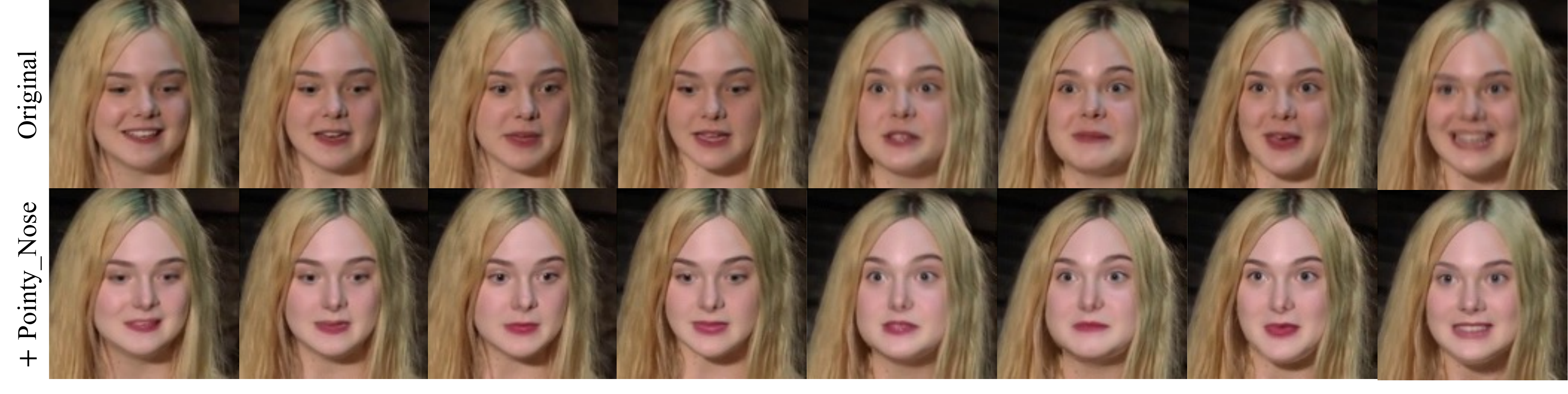}
  \includegraphics[width=0.98\linewidth]{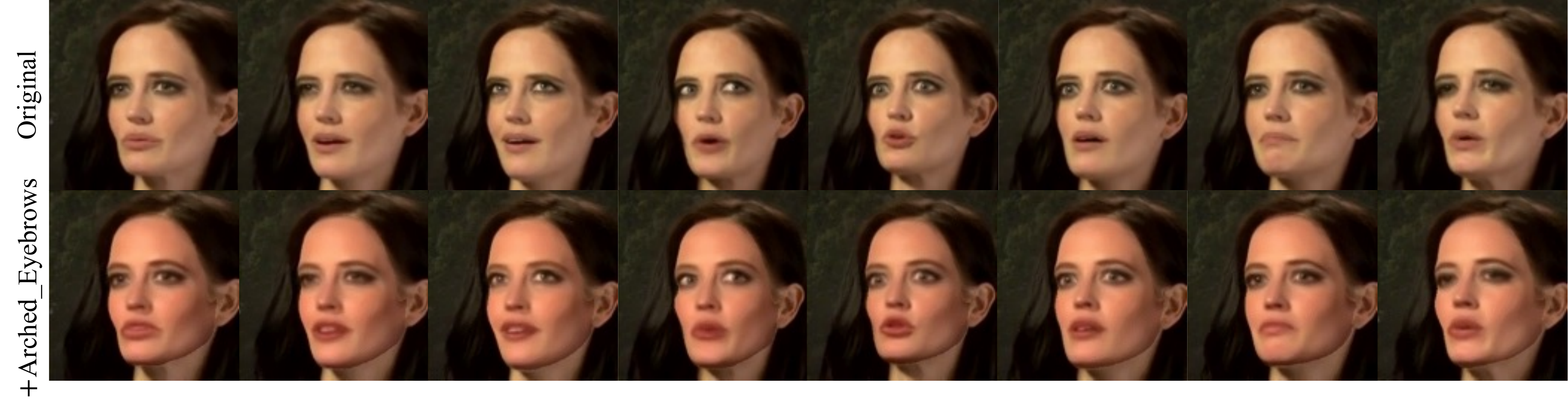}
  \caption{Classifier-based video editing on VoxCeleb1 test set.}
  \label{fig:add-classifier-vox}
\end{figure*}

\begin{figure*}
  \centering
  \includegraphics[width=0.98\linewidth]{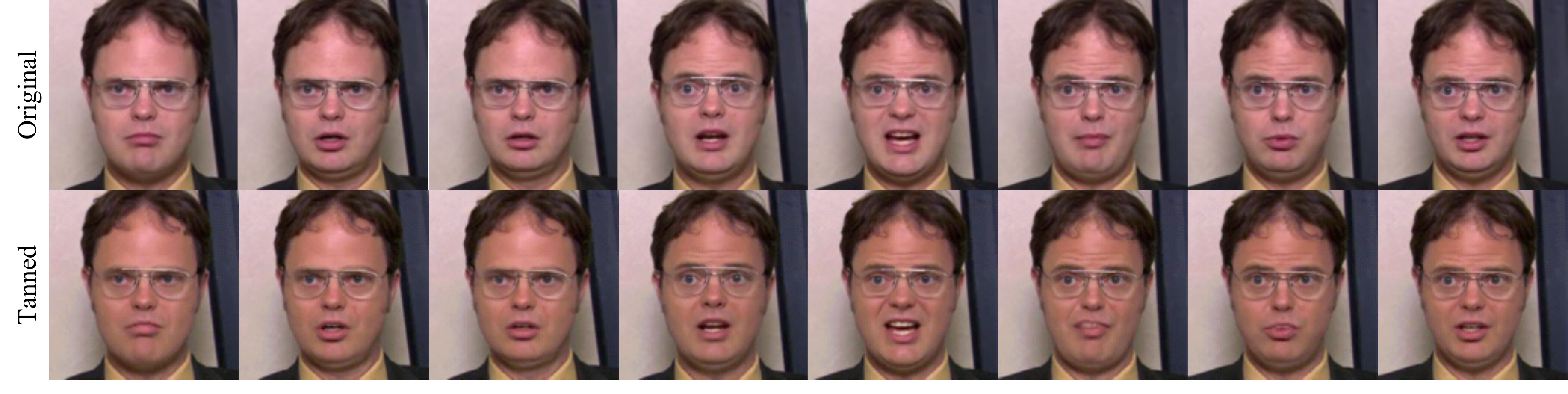}
  \includegraphics[width=0.98\linewidth]{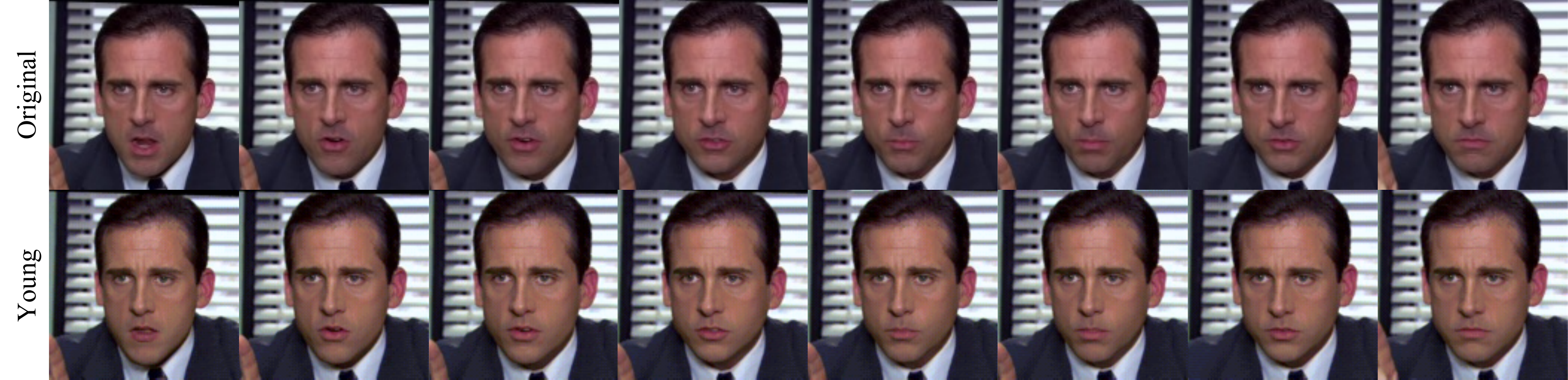}
  \caption{CLIP-based video editing on the other videos not in VoxCeleb1.}
  \label{fig:add-clip-stit}
\end{figure*}

\begin{figure*}
  \centering
  \includegraphics[width=0.98\linewidth]{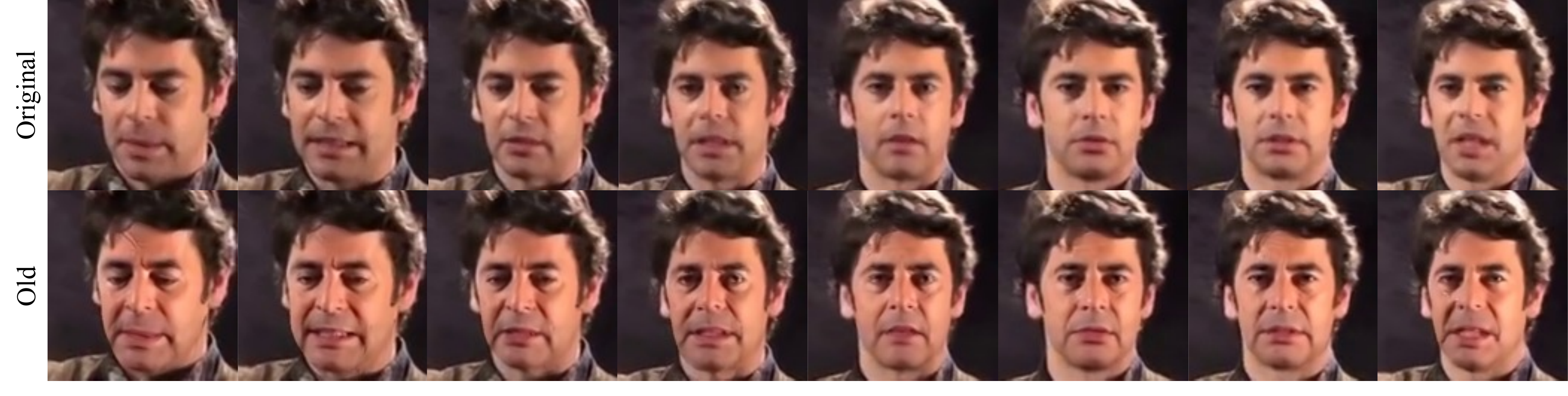}
  \includegraphics[width=0.98\linewidth]{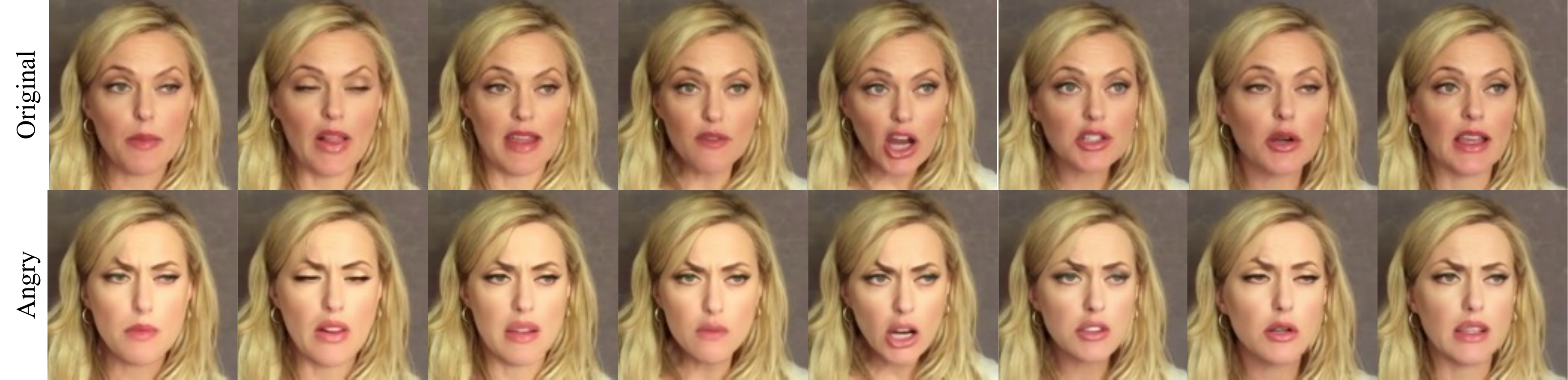}
  \caption{CLIP-based video editing on VoxCeleb1 test set.}
  \label{fig:add-clip-vox}
\end{figure*}

We show additional video editing results with classifier-based editing in \cref{fig:add-classifier-stit,fig:add-classifier-vox} and CLIP-based editing in \cref{fig:add-clip-stit,fig:add-clip-vox}. These results demonstrate that our video editing method has temporal consistency for other attributes as well.

\end{document}